%% file: main.tex
\documentclass{article} %
\usepackage{conference,times}

\input{math_commands.tex}

\input{shortcut}

\usepackage{hyperref}
\usepackage{url}
\usepackage{graphicx}
\usepackage{algorithm}
\usepackage{algpseudocode}
\usepackage{multirow}
\usepackage{amsmath}
\usepackage{amsfonts}
\usepackage{nicematrix}
\usepackage{enumitem}
\usepackage{booktabs, tabularx}
\usepackage{threeparttable}
\usepackage{bbding}
\usepackage{utfsym}
\usepackage{pifont}
\usepackage{mathrsfs}

\makeatletter
\@ifundefined{Cross}{}
  {}
\makeatother
\usepackage{marvosym}

\title{3DTrajMaster: Mastering 3D Trajectory for Multi-Entity Motion in Video Generation}

\author{
Xiao Fu$^{1}$ \;
Xian Liu$^{1}$ \;
Xintao Wang$^{2}{\textsuperscript{\Letter}}$ \;
Sida Peng$^{3}$ \;
Menghan Xia$^2$ \;
Xiaoyu Shi$^2$ \\
\textbf{Ziyang Yuan}$^2$ \;
\textbf{Pengfei Wan}$^{2}$ \; 
\textbf{Di Zhang}$^2$ \;
\textbf{Dahua Lin}$^{1}{\textsuperscript{\Letter}}$\\
$^1$The Chinese University of Hong Kong \; 
$^2$Kuaishou Technology \; 
$^3$Zhejiang University 
}

\usepackage{color}
  
\definecolor{red}{RGB}{255,0,0}

\newcommand\nnfootnote[1]{%
  \begin{NoHyper}
  \renewcommand\thefootnote{}\footnote{#1}%
  \addtocounter{footnote}{-1}%
  \end{NoHyper}
}

\iclrfinalcopy %

\begin{document}

\maketitle

\input{main/abstract}
\input{main/introduction}
\input{main/relatedwork}

\input{main/method}
\input{main/experiments}

\input{main/conclusion}
\input{main/acknowledgment}

\bibliography{conference}
\bibliographystyle{conference}

\newpage
\input{supp/supp}

\end{document}

%% file: math_commands.tex
\usepackage{amsmath,amsfonts,bm}

\def\eqref#1{equation~\ref{#1}}

\def\1{\bm{1}}

\DeclareMathAlphabet{\mathsfit}{\encodingdefault}{\sfdefault}{m}{sl}
\SetMathAlphabet{\mathsfit}{bold}{\encodingdefault}{\sfdefault}{bx}{n}

%% file: shortcut.tex
\newcommand{\bK}{\mathbf{K}}

\newcommand{\bk}{\mathbf{k}}
\newcommand{\bx}{\mathbf{x}}

\newcommand{\bV}{\mathbf{V}}

\newcommand{\bv}{\mathbf{v}}
\newcommand{\bq}{\mathbf{q}}

\newcommand{\bT}{\mathbf{T}}
\newcommand{\bQ}{\mathbf{Q}}

\makeatletter
\DeclareRobustCommand\onedot{\futurelet\@let@token\@onedot}
\def\@onedot{\ifx\@let@token.\else.\null\fi\xspace}
\def\eg{e.g.}

\makeatother

\newcommand{\figref}[1]{Fig.~\ref{#1}}
\newcommand{\secref}[1]{Sec.~\ref{#1}}

\renewcommand{\eqref}[1]{Eq.~\ref{#1}}
\newcommand{\tabref}[1]{Table~\ref{#1}}

\newif\ifcomment
\commenttrue
\ifcomment
	\newcommand{\yl}[1]{ \noindent {\color{cyan} {\bf Yiyi:} {#1}} }

\else
	\newcommand{\ag}[1]{}
	\newcommand{\yl}[1]{}
\fi

%% file: main/abstract.tex
\begin{abstract}

This paper aims to manipulate multi-entity 3D motions in video generation. Previous methods on controllable video generation primarily leverage 2D control signals to manipulate object motions and have achieved remarkable synthesis results. However, 2D control signals are inherently limited in expressing the 3D nature of object motions. To overcome this problem, we introduce \textbf{3DTrajMaster}, a robust controller that regulates multi-entity dynamics in \textit{3D space}, given user-desired 6DoF pose (location and rotation) sequences of entities. At the core of our approach is a plug-and-play 3D-motion grounded object injector that fuses multiple input entities with their respective 3D trajectories through a gated self-attention mechanism. In addition, we exploit an injector architecture to preserve the video diffusion prior, which is crucial for generalization ability.
To mitigate video quality degradation, we introduce a domain adaptor during training and employ an annealed sampling strategy during inference. To address the lack of suitable training data, we construct a 360$^{\circ}$-Motion Dataset, which first correlates collected 3D human and animal assets with GPT-generated trajectory and then captures their motion with 12 evenly-surround cameras on diverse 3D UE platforms. Extensive experiments show that 3DTrajMaster sets a new state-of-the-art in both accuracy and generalization for controlling multi-entity 3D motions. Project page: \url{http://fuxiao0719.github.io/projects/3dtrajmaster}.

\nnfootnote{\textsuperscript{\Letter}: Corresponding Authors.}

\end{abstract}

%% file: main/introduction.tex
\section{Introduction}

\begin{figure*}[ht]
\centering
\includegraphics[width=.95\linewidth]{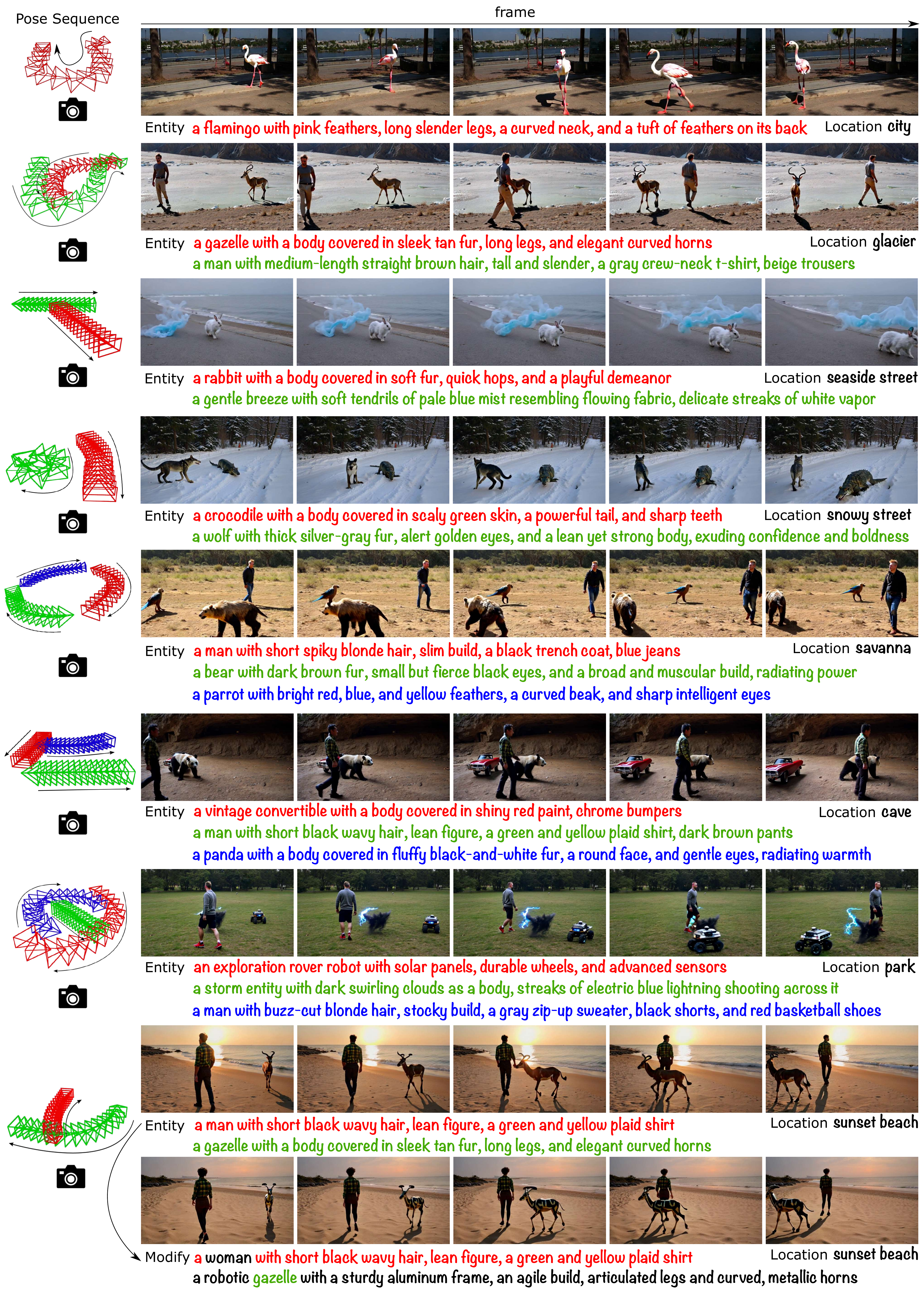}
\caption{\textbf{3DTrajMaster controls one or multiple entity motions in 3D space with input entity-specific 3D trajectories for text-to-video (T2V) generation.} It allows diverse entity categories (human, animal, car, robot, natural force, etc) and flexible edits on entity descriptions (see more in~\figref{fig:supp_diffhuman}). The text prompt is ``\textit{\{Entity 1\},..., and \{Entity N\} is/are moving in the \{Location\}}". \textcolor{red}{\textit{(We kindly urge readers to check more generalizable results ($\geq$200) in our website)}}}
\label{fig:teaser}

\vspace{-2ex}
\end{figure*}

Controllable video generation~\citep{videoworldsimulators2024,guo2023animatediff,chen2023videocrafter1} aims to synthesize high-fidelity videos that are controlled by user inputs, such as text prompts, sketches, or bounding boxes.
A critical objective in controllable video generation is the precise manipulation of object motions within videos, which is essential for simulating the dynamic world and potentially aids video generative models in understanding the underlying physics of the world.
In addition, it can unleash many applications of video generative models, such as virtual cinematography for the film industry, acting as interactive games, and providing world models for embodied AI systems.

Recently, there has been some methods attempting to manipulate object motions in video generation by introducing 2D control signals, such as 2D sketches~\citep{wang2024videocomposer,guo2023sparsectrl}, bounding boxes~\citep{yang2024direct,wang2024boximator}, and points~\citep{wang2024motionctrl,zhang2024tora}.
These methods offer convenient user interactions and have delivered impressive video generation results.
However, we argue that 2D control signals cannot fully express the inherent 3D nature of motion, which limits the control capability of object motions.
As real-world objects move in 3D space, some motion properties can only be described through 3D representations.
For example, the rotation of an object can be succinctly described using three parameters in 3D, and occlusions between objects can be simply represented using z-buffering. 
In contrast, it is quite difficult for 2D control signals to represent these concepts.

In this paper, we focus on the problem of controlling multi-object 3D motions in video generative models, aiming to simulate the authentic dynamics of objects in 3D space.
This setting is more aligned with the requirements of downstream applications, such as emulating realistic human motions in movies or exploring 3D virtual scenes in games.
However, this problem is extremely challenging.
There are three core questions we need to answer:
1) How to precisely represent the 3D motions of objects; 
2) How to correlate multiple object descriptions with their respective motion sequences in video generative models;
3) How to maintain the generalization capability of video models after injecting 3D motion information.

To address these, we propose a novel approach, \textbf{3DTrajMaster}, which is able to manipulate multi-entity motions in 3D space for video generation by leveraging entity-specific 6DoF pose sequences as additional inputs.
The core of our model is a plug-and-play 3D-motion grounded object injector, which associates each entity with their corresponding pose sequences, and then injects these conditions into the foundation model, to control the entity motion. Specifically, the entities and trajectories are projected into latent embeddings via a frozen text encoder and a learnable pose encoder, respectively. 
These two modality embeddings are then entity-wise added to form correspondences, which are further fed into a gated self-attention layer for motion fusion. This plug-and-play architecture preserves the video model’s prior and can generalize on more diverse entities and 3D trajectories.

However, another challenge in training our model lies in data availability. Existing video datasets face two key limitations: 
1) \textit{Low entity diversity:} Datasets with paired entities and 3D trajectories are mostly limited to humans and autonomous vehicles, with inconsistent spatial distributions and overcrowded entities. 
2) \textit{Inaccurate/Failed pose estimation:} Current 6D pose estimation methods focus on rigid objects, while non-rigid objects, such as animals, are underrepresented, with only human poses studied using SMPL~\citep{loper2023smpl}. 
To this end, we choose to construct a custom dataset, termed 360$^{\circ}$-Motion Dataset, with unified trajectory distribution using advanced UE rendering techniques. 
We start by collecting 3D assets of humans and animals and rescaling them to a unified cubic space. 
GPT~\citep{achiam2023gpt} is then employed to generate 3D trajectory templates for these assets. 
Various entities and trajectory templates are arranged and combined to create diverse motions. These globally animated assets are captured using 12 evenly positioned cameras within the collected 3D scenes, including city (MatrixCity~\citep{li2023matrixcity}), desert, forest, and HDRIs (projected into 3D space)\footnote{Poly Haven: https://polyhaven.com/}. To prevent video domain shift in our constructed dataset, we introduce two key components: 
1) A video domain adaptor, which is trained to fit data distribution and slightly reduced during inference. 
2) An annealed sampling strategy, where trajectories are injected to guide general motion in the early steps and drop out in the later stages. 

We evaluate our 3DTrajMaster in the curated novel pose sequences with GPT-generated entity prompts, obtaining a significant lead over current SOTAs. In summary, our contributions are:

1) We are the first to customize 6 degrees of freedom (DoF) multi-entity motion in 3D space for controllable video generation, establishing a new benchmark for fine-grained motion control.

2) We propose a 3D-motion grounded video diffusion model that controls multi-entity motions using pose sequences as motion representations. Our flexible object injector enforces entity-wise correspondence between objects and their motions and preserves the video diffusion prior.

3) We introduce a scalable 4D motion dataset construction mechanism, and techniques like the video domain adaptor and annealed sampling to enhance video quality while maintaining motion accuracy.

4) 3DTrajMaster achieves state-of-the-art accuracy in controlling 3D entity motions and allows fine-grained entity input customization such as changing human hair, clothing, gender, and figure size.

%% file: main/relatedwork.tex
\section{Related Work}

\noindent \textbf{Customizing Video Motion with 2D Guidance.} Previous methods predominantly perform motion control on 2D spaces, as this aligns more easily with the input video format. A straightforward path is to direct videos based on motion patterns from reference videos~\citep{zhao2023motiondirector,jeong2024vmc,ling2024motionclone}. However, they require users to provide reference video templates.
While training-free paradigms~\citep{yang2024direct,xiao2024video}, utilizing attention mechanisms to edit spatial-temporal layouts, can mitigate this issue, they exhibit poor generalization in real-world scenarios and rely heavily on trial-and-error. Further advancements utilize more high-level representations, such as sketches\&depths (dense or sparse)~\citep{wang2024videocomposer,guo2023sparsectrl}, pose skeletons~\citep{feng2023dreamoving,xu2024magicanimate,chen2024dreamcinema}, bounding boxes~\citep{wang2024boximator}, and 2D trajectories~\citep{wang2024motionctrl,zhang2024tora,yin2023dragnuwa,yang2024direct}, to enable more flexible motion generation. Although these methods can model camera, object, or joint movements, the lack of 3D awareness limits precise 3D motion control.

\noindent \textbf{Learning 3D-aware Motion Synthesis.} Considering that video is a sequence of images projected from 3D world, manipulating video in 3D space is both more crucial and impactful. A key aspect of this manipulation is camera movement. MotionCtrl~\citep{wang2024motionctrl} is the first to regulate video using camera poses (rotation and translation) in 3D space, while CameraCtrl~\citep{he2024cameractrl} and VD3D~\citep{bahmani2024vd3d} further enhance camera representation with plücker embeddings~\citep{sitzmann2021light}. SynCamMaster~\citep{bai2024syncammaster} extends single-camera control to multi-camera synchronization. GameGen-X~\citep{che2024gamegen} can generate game videos with novel `WASD' keyboard inputs. Other approaches~\citep{hou2024training, hu2024motionmaster} also explore training-free paradigms. However, none address the customization of object motion in 3D space. Manipulation on 2D maps~\citep{wang2024motionctrl,zhang2024tora} often fails in multi-object scenarios, particularly with 1) aligning each entity and its corresponding motion, 2) handling \textit{3D occlusion}. In contrast, 3DTrajMaster is the first to overcome them and simulate plausible 3D motions.

%% file: main/method.tex
\section{3DTrajMaster}

\begin{figure*}[h]
\centering
\includegraphics[width=\linewidth]{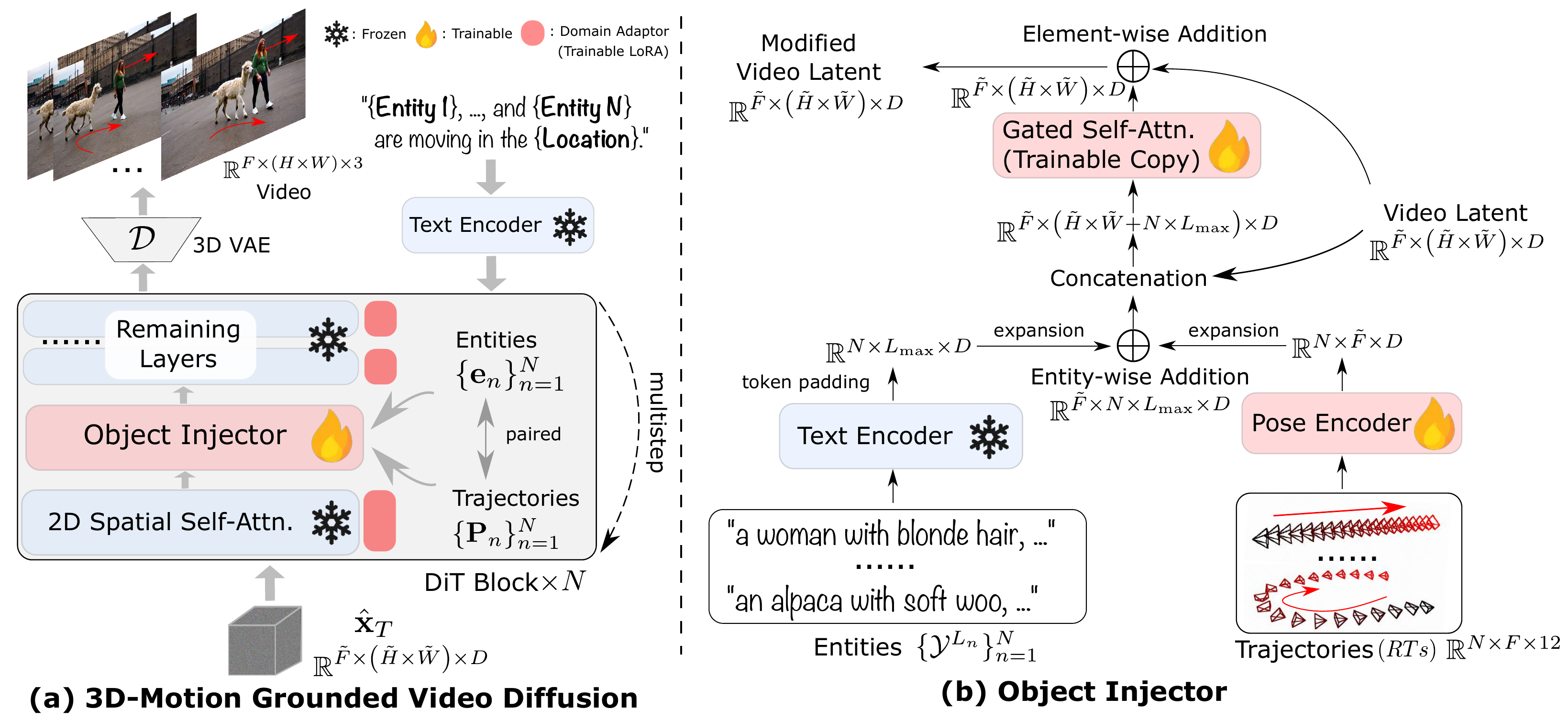}
\vspace{-1em}
\caption{\textbf{3DTrajMaster Framework.} Given a text prompt consisting of N entities $\{\mathbf{e}_n\}_{n=1}^N$, 3DTrajMaster (a) is able to generate the desired video with entity motions that conform to the input entity-wise pose sequences $\{\mathbf{P}_n\}_{n=1}^N$. 
Specifically, it involves two training phases.  
First, it utilizes a domain adaptor to mitigate the negative impact of training videos. Then, an object injector module is inserted after the 2D spatial self-attention layer to integrate paired entity prompts and 3D trajectories. 
(b) Details of the object injection process. 
The entities are projected into latent embeddings through the text encoder. The paired pose sequences are projected using a learnable pose encoder and then fused with entity embeddings to form entity-trajectory correspondences. 
This condition embedding is concatenated with the video latent and fed into a gated self-attention layer for motion fusion. 
Finally, the modified latent gets back to the remaining layers in the DiT block.
}
\label{fig:pipeline}
\vspace{-1em}
\end{figure*}

Our goal is to master entity motions in 3D space for text-to-video (T2V) generation by leveraging entity-specific 3D trajectories as additional inputs. To this end, we introduce \textit{3DTrajMaster} (see~\figref{fig:pipeline}), a 3D-motion grounded video diffusion model trained in two stages. First, we describe the video diffusion model and the task formulation (\secref{sec:preliminaries}). Then, we present our proposed model, whose core is to train a plug-and-play 3D grounded object injector to integrate multiple detailed entity descriptions and the respective pose sequences (\secref{sec:3d_grounded_object_injector}). We further incorporate a domain adaptor to mitigate video domain shifts introduced by our constructed training data (\secref{sec:lora_module}). Finally, we detail the inference process using annealed sampling to enhance video quality (\secref{sec:inference}). 

\subsection{Preliminaries on 3D-Entity-Aware Video Distribution} \label{sec:preliminaries}

\noindent \textbf{Video Diffusion Models.} Latent text-to-video diffusion model~\citep{ho2022imagen,ho2022video,videoworldsimulators2024,chen2023videocrafter1,blattmann2023stable} learns the conditional distribution $p(\mathbf{x}|\mathbf{c})$ of encoded video data $\mathbf{x}$ ($\mathbf{x}=\mathcal{E}(X)$, $\mathcal{E}(\cdot)$ is VAE encoder) given text description $\mathbf{c}$ in latent space. In the forward progress, it progressively transits the clean data $\mathbf{x}_{0}$ to the desired Gaussian distribution in a Markov chain: $\left\{\mathbf{x}_t, t \in(1, T) \mid \mathbf{x}_t=\alpha_t \mathbf{x}_0+\sigma_t \boldsymbol{\epsilon}, \boldsymbol{\epsilon} \sim \mathcal{N}(\mathbf{0}, \mathbf{I}) \right\}$. 
To iteratively recover the data $\hat{\mathbf{x}}_{0}$ from the noise $\boldsymbol{\epsilon} \sim \mathcal{N}\left(\mathbf{0}, \sigma_{t}^2 \mathbf{I}\right)$, it learns a denoising model $\hat{\boldsymbol{\epsilon}}_{\boldsymbol{\theta}}$ with the objective function: $\boldsymbol{\epsilon} \approx \hat{\boldsymbol{\epsilon}}_{\boldsymbol{\theta}} \left(\mathbf{x}_t;t,\mathbf{c}\right)$. 
With the preconditioning strategy~\citep{karras2022elucidating,salimans2022progressive}, it optimizes the neural network $\hat{F}_{\boldsymbol{\theta}}$ by parameterizing the $\hat{\boldsymbol{\epsilon}}_{\boldsymbol{\theta}}$ as: $\hat{\boldsymbol{\epsilon}}_{\boldsymbol{\theta}} = c_{\text{out}}(\sigma_t) \hat{F}_{\boldsymbol{\theta}}\left(c_{\text {in}}(\sigma_t) \mathbf{x}_t ; \boldsymbol{c}, \sigma_t\right)+c_{\text {skip }}(\sigma_t) \mathbf{x}_t$. 

\noindent \textbf{Task Formulation.} Given an input text prompt $\boldsymbol{c}$ consisting of N entities $\{\mathbf{e}_n\}_{n=1}^N$ and their paired 3D trajectories $\{\mathbf{P}_n\}_{n=1}^N$, where $\mathbf{P}_n^f=[\mathbf{R}; \mathbf{T}] \in \mathbb{R}^{3 \times 4}$ for $f$-th frame and object orientation and translation are represented by $\mathbf{R} \in \mathbb{R}^{3\times3}$ and $\mathbf{T} \in \mathbb{R}^{3}$, respectively, our goal is to generate plausible video $\mathbf{X} \in \mathbb{R}^{F \times H \times W}$ that accords with each entity description $\mathbf{e}$ and the respective trajectory $\mathbf{P}$. The overall generative formulation $f(\cdot)$ is 
\begin{equation}
f(\cdot): \mathbf{c} \in \mathcal{Y}^L, (\mathbf{e}_n \in \mathcal{Y}^{L_n},\mathbf{P}_n \in \mathbb{R}^{3 \times 4})_{n=1}^N \rightarrow \mathbf{X} \in \mathbb{R}^{F \times H \times W} 
\end{equation}
where $\mathbf{X} \approx \mathcal{D}(\hat{\mathbf{x}}_0)$ ($\mathcal{D}(\cdot)$ is the VAE decoder), $\hat{\mathbf{x}} = p\left(\hat{\mathbf{x}}_T \right) \prod_{t=1}^T p_{\boldsymbol{\theta}}\left(\hat{\mathbf{x}}_{t-1} \mid \hat{\mathbf{x}}_t, \mathbf{c}, (\mathbf{e}_n,\mathbf{P}_n)_{n=1}^N) \right)$, $\mathcal{Y}$ is the alphabet, and $L$ is the token length. Our primary challenge lies in modeling the distribution $p_{\boldsymbol{\theta}}$ or specifically $\hat{\boldsymbol{\epsilon}}_{\boldsymbol{\theta}}$ to generate realistic videos that accurately correspond to the given multiple 3D entity conditions. Here we structure $\hat{\boldsymbol{\epsilon}}_{\boldsymbol{\theta}}(\mathbf{x}; \boldsymbol{c}, \sigma_t, (\mathbf{e}_n,\mathbf{P}_n)_{n=1}^N)$ as transformer architecture~\citep{peebles2023scalable} for its superior scalability and performance over U-Net~\citep{ronneberger2015u}.

\subsection{Plug-and-play 3D-Motion Grounded Object Injector} \label{sec:3d_grounded_object_injector}

\noindent \textbf{Matching Entity-Trajectory Pair.} The entity prompts $\{\mathbf{e}_n\}_{n=1}^N$ are projected into latent embeddings $\{\mathbf{Z}^{\mathbf{e}}_n\}_{n=1}^N$ using a frozen text encoder $\mathcal{E}_{\mathbf{T}}(\cdot): \mathbf{e}_n \in \mathcal{Y}^{L_n} \rightarrow \mathbf{Z}^{\mathbf{e}}_n \in \mathbb{R}^{L_{\text{max}}\times D}$, where each embedding $\mathbf{Z}^{\mathbf{e}}_n$ is zero-padded to maximum token length $L_{\text{max}}$. Correspondingly, the pose sequences $\{\mathbf{P}_n\}_{n=1}^N$ are also projected into latent embeddings $\{\mathbf{Z}^{\mathbf{P}}_n\}_{n=1}^N$ through the trainable pose encoder $\mathcal{E}_{\mathbf{P}}(\cdot)$: $\mathbf{P}_n \in \mathbb{R}^{F \times 12} \rightarrow \mathbf{Z}^{\mathbf{P}}_n \in \mathbb{R}^{\tilde{F} \times D}$. The pose encoder $\mathcal{E}_{\mathbf{P}}$ consists of a linear layer and a downsampler along the temporal dimension, resembling the causal encoding applied to video input $\mathbf{x}$ in 3D VAE, where the mapping function is $\mathcal{E}_{\mathbf{X}}(\cdot)$: $\mathbf{X} \in \mathbb{R}^{F \times H \times W} \rightarrow \mathbf{x}\in \mathbb{R}^{\tilde{F} \times \tilde{H} \times \tilde{W}}$. Here the downsampler refers to interval sampling of tensors, where we also tried several sequential one-dimensional convolution layers but achieved similar results. Then, the paired entity and trajectory embeddings are expanded and combined through entity-wise addition to form a bonded entity-motion correspondence $\mathbf{Z}^\mathbf{Pe}\in \mathbb{R}^{\tilde{F} \times N \times L_{\text{max}} \times D}$.

\noindent \textbf{Gated Self-Attention for Motion Fusion.} Inspired by \citep{li2023gligen}, we employ a gated self-attention layer to handle multiple entity-trajectory pairs $\mathbf{Z}^\mathbf{Pe}$ (with varying dimensional embeddings) as input, while further refining the correlated features. Specifically, we replicate the weight of the 2D spatial self-attention layer in each DiT block as initialization to enable grounding. The input video tokens $\mathbf{x}_t$ and $\mathbf{Z}^\mathbf{Pe}$ are passed through this trainable copy via truncated self-attention. The output can be expressed in a residue-connection form:
\begin{equation}
\begin{split}
& \mathbf{x}_t = \mathbf{x}_t + \beta \cdot \mathbf{Tc}(\mathbf{Att}(\mathbf{q}, \mathbf{k}, \mathbf{v})) \\
\bq =\bQ \cdot \bT&,  \bk =\bK \cdot \bT, \bv=\bV \cdot \bT, \bT=\mathbf{x}_t \oplus \mathbf{Z}^\mathbf{Pe}
\end{split}
\end{equation}
where $\beta$ is a trainable scale, $\mathbf{Tc}(\cdot)$ is the truncation operation to preserve $\mathbf{x}_n$ tokens, $\mathbf{Att(\cdot)}$ is softmax attention, $\mathbf{Q}$, $\mathbf{K}$ and $\mathbf{V}$ are query, key and value embedding matrices, and $\oplus$ denotes concatenation. In this stage, we train the $\boldsymbol{\theta_1}$ including the pose encoder and the gated self-attention parameters as follow.
\begin{equation}
\mathcal{L}(\boldsymbol{\theta_1})=\mathbb{E}_{\mathbf{x}, \mathbf{c}, \boldsymbol{\epsilon} \sim \mathcal{N}\left(\mathbf{0}, \sigma_{t}^2 \mathbf{I}\right), \mathbf{e}, \mathbf{P}, t, \beta}\left[\left\|\boldsymbol{\epsilon}-\hat{\boldsymbol{\epsilon}}_{\boldsymbol{\theta}_1}\left(\mathbf{x}_t, \mathbf{c}, (\mathbf{e}_n,\mathbf{P}_n)_{n=1}^N), t, \beta \right)\right\|_2^2\right]
\end{equation}

\subsection{Alleviating Video Domain Shift from Constructed Training Data} \label{sec:lora_module}

\begin{figure*}[ht]
\centering
\includegraphics[width=\linewidth]{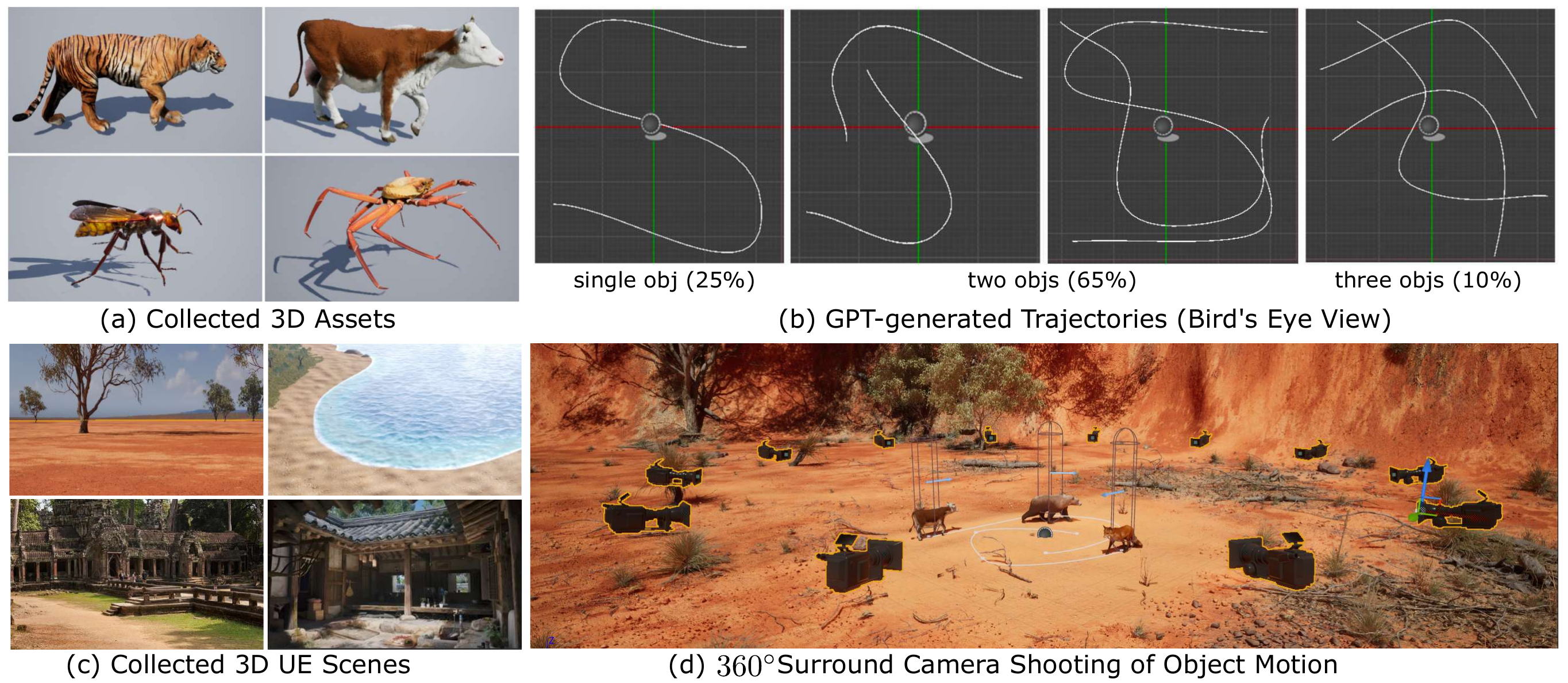}
\vspace{-2em}
\caption{\textbf{Dataset Construction Illustration.}
We correlate (a) collected 3D assets with (b) GPT-generated 3D trajectories on (c) diverse 3D UE platforms, positioning (d) 12 evenly distributed surrounding cameras to capture the object motions in video format.
}
\label{fig:dataset}
\vspace{-1em}
\end{figure*}

\noindent \textbf{360$^{\circ}$-Motion Dataset.} 
High-quality training data is vital for learning generalizable 3D motion control. A straightforward preparation is to extract paired entity descriptions and 6DoF poses from common video datasets. However, it is hard due to twofold: 1) \textit{Low diversity/quality entity}:  Datasets with paired entities and 3D trajectories are mostly limited to humans~\citep{jiang2024scaling,araujo2023circle} and autonomous vehicles~\citep{geiger2012we,sun2020scalability}, where the spatial distributions vary between datasets and the entities may be overcrowded. In video datasets like Artgrid, Pixabay, and Pexels\footnote{Artgrid: https://artgrid.io/, Pixabay: https://www.videvo.net/, Pexels: https://www.pexels.com/}, human category occupies a relatively large proportion in 3D/4D asset objectives (refer to~\secref{subsec:unbalanceddataset}), limiting model generalization to other categories like animals and vehicles. Issues like watermarks in WebVid~\citep{bain2021frozen} further increase the cost of filtering. 2) \textit{Low-accuracy/Failed pose estimation}: Most 6D pose estimation methods exclusively focus on rigid objects, and rely on CAD models~\citep{labbe2022megapose,wen2024foundationpose} or posed multi-view images~\citep{liu2022gen6d,sun2022onepose}. For non-rigid animated objects, only human poses have been widely studied via methods like SMPL~\citep{loper2023smpl}, limiting the estimation for general 4D objects, such as animals. A simpler alternative is to represent only 3D locations via depth models~\citep{hu2024depthcrafter,ke2024video,fu2025geowizard}. However, there exist errors in segmenting the foreground entities from the background and can not generate consistent video metric depth.

To circumvent the aforementioned challenges, we opt to construct a synthetic dataset, named \textit{360$^{\circ}$-Motion}, through Unreal Engine (UE) with advanced rendering technologies (see~\figref{fig:dataset}). We begin by collecting 70 animated 3D assets across two categories: human and animal. Humans are differentiated by attributes such as gender, clothing, body shape, and hairstyle. GPT-4V~\citep{openai2023gpt4v} is then used to generate text descriptions $\mathbf{e}_n \in \mathcal{Y}^{L_n} ({L_n} \leq 20)$ for each rendered asset image (\figref{fig:dataset} (a)). For posed object trajectory templates (\figref{fig:dataset} (b)), we follow TC4D~\citep{bahmani2024tc4d} by leveraging GPT to generate 3D spline (location $\mathbf{T}$) and additional orientation $\mathbf{R}$ via the gradient calculation on spline. This process yields approximately 96 templates in canonical space, each associated with one to three assets. We additionally reduce the size of the animals by a ratio of 0.6 to prevent collisions with other assets. The paired assets and their motion templates are then placed within a 5$\times$5 square meter range in one of the 3D platforms, including city (MatrixCity~\citep{li2023matrixcity}), dessert, forest, and HDRIs (projected into 3D). We position 12 sets of cameras evenly around the scene to capture 360-degree views, producing 100 frames per video clip at 384$\times$672 resolution for each camera. This process produces a total of 54,000 videos by arranging and combining various objects and trajectories. (see~\secref{subsec:360motiondataset} and Supp. video samples for illustration)

\noindent \textbf{Video Domain Adaptor.} Training video diffusion models on this relatively small set of constructed video clips can lead to an undesirable UE style, limiting the generalization ability. To prevent learning this variation in quality and retain the knowledge of the base T2V, we train LoRA modules~\citep{hu2021lora} that serve as video domain adaptor. Specifically, we integrate LoRA into self-attention, cross-attention, and linear layers of the base T2V model, as shown in~\figref{fig:pipeline}. The attention/linear projection matrices $\{\mathbf{W}_n\}_{n=1}^K$ are associated with additional trainable lower rank matrices $\{\Delta\mathbf{W}_n=\alpha \mathbf{A}_n \mathbf{B}_n^T\}_{n=1}^K$, where $\alpha$ is the scaler that can be adjusted to control the adaptor influence. During inference, we set $\alpha$ to a small value to mitigate the negative impact of synthetic video data. We optimize $\boldsymbol{\theta_2}=\{\Delta\mathbf{W}_n\}_{n=1}^K$ with the training objective:

\begin{equation}
\mathcal{L}(\boldsymbol{\theta_2})=\mathbb{E}_{\mathbf{x}, \mathbf{c}, \boldsymbol{\epsilon} \sim \mathcal{N}\left(\mathbf{0}, \sigma_{t}^2 \mathbf{I}\right), t, }\left[\left\|\boldsymbol{\epsilon}-\hat{\boldsymbol{\epsilon}}_{\boldsymbol{\theta}_1}\left(\mathbf{x}_t, \mathbf{c}, t, \alpha \right)\right\|_2^2\right].
\end{equation}

Note that the domain adaptor $\boldsymbol{\theta_2}$ is frozen when training the object injector $\boldsymbol{\theta_1}$. 

\subsection{Inference Procedure} \label{sec:inference}
We initialize the video latent $\hat{\bx}_T$ as standard Gaussian noise, and progressively denoise it with the guidance of desired entity-trajectory pairs $(\mathbf{e}_n,\mathbf{P}_n)_{n=1}^N$, following the same schedule as the previous two training stages. We apply classifier-free guidance~\citep{ho2022classifier} and use DDIM~\citep{song2020denoising} for re-spaced sampling for acceleration. To further enhance the video quality, we employ an annealed sampling strategy (Algorithm 1): During inference in the former steps, trajectories are inserted into the model to define the general object motions, while in the latter stage, they are dropped out, transitioning to the standard T2V generation process. We also observe that setting negative 3D trajectories as static motions $\{(\hat{\mathbf{P}}_n)_{n=1}^N|\hat{\mathbf{P}}_n=\mathbf{P}_0,\forall n\}$ can further improve pose accuracy. This phenomenon reflects the model's ability to learn 3D motion representations: Since we do not randomly drop out motion sequences during training like text, the model implicitly learns static motion modeling from videos where entities are primarily in motion. Thus when setting static motion as a ``negative motion prompt”, we can amplify the magnitude of entity movement, leading to improved pose accuracy during evaluation. However, we do not adopt it as it sometimes results in a video quality decline (refer to~\secref{subsec:negpose}).

\begin{algorithm} 
\caption{Annealed conditional sampling with classifier-free guidance (CFG)} \label{alg:infer}
\begin{algorithmic}[1]
\Require $w$: guidance strength, $T_c$: annealed timestep, $\alpha$: LoRA modulator, $\tilde{\boldsymbol{\theta}}$: frozen base T2V model, $\boldsymbol{\theta}_1$: object injector, $\boldsymbol{\theta}_2$: domain adaptor, $\mathbf{c}$: text condition, $(\mathbf{e},\mathbf{P})$: entity-trajectory pairs  \\
$\hat{\mathbf{x}}_1 \sim \mathcal{N}\left(\mathbf{0}, \sigma_{t}^2 \mathbf{I}\right)$
\For{$t=1,...,T$}
\If{$\leq T_c$} 
    \State $\tilde{\epsilon}_t=(1+w) \hat{\boldsymbol{\epsilon}}_{\tilde{\boldsymbol\theta},\boldsymbol\theta_1,\boldsymbol\theta_2}\left(\hat{\mathbf{x}}_t, \mathbf{c}, (\mathbf{e}_n,\mathbf{P}_n)_{n=1}^N, \alpha\right)-w \hat{\boldsymbol{\epsilon}}_{\tilde{\boldsymbol\theta},\boldsymbol\theta_1,\boldsymbol\theta_2}\left(\hat{\mathbf{x}}_t, \alpha\right)$
\Else{}
    \State $\hat{\epsilon}_t=(1+w) \boldsymbol{\epsilon}_{\tilde{\boldsymbol\theta}}\left(\hat{\mathbf{x}}_t, \mathbf{c}\right)-w\boldsymbol{\epsilon}_{\tilde{\boldsymbol\theta}}\left(\hat{\mathbf{x}}_t\right)$ 
\EndIf 
\State $\hat{\boldsymbol{z}}_t = \left(\hat{\mathbf{x}}_t-\sigma_t \tilde{\boldsymbol{\epsilon}}_t\right) / \alpha_t$
\State $\hat{\mathbf{x}}_{t+1} \sim \mathcal{N}\left(\hat{\mathbf{x}}_{t+1} ; \tilde{\boldsymbol{\mu}}_{t+1 \mid t}\left(\hat{\boldsymbol{z}}_t, \hat{\mathbf{x}}_t\right), \sigma_{t+1 \mid t}^2 \mathbf{I}\right)$ if $t < T$ else $\hat{\mathbf{x}}_{t+1}=\hat{\boldsymbol{z}}_t$
\EndFor
\State \Return $\hat{\mathbf{x}}_{t+1}$
\end{algorithmic}
\end{algorithm}

\vspace{-1.5em}

%% file: main/experiments.tex
\section{Experiments}
\subsection{Implementation Details}
For input text prompts, we use a unified template: ``\textit{\{Entity 1\},..., and \{Entity N\} are moving in the \{Location\}.}" Here we set ``\{Location\}"  based on the respective 3D UE platform. We train 3DTrajMaster based on our internal video diffusion model for research purposes (see~\secref{subsec:videobackbone} for more details), which contains $\sim$ 1B parameters. The clipped training video and inference video are set to $384\times672$ resolutions. Each video segment is 5 seconds long. We utilize the Adam optimizer and train on a cluster of 8 NVIDIA H800 GPUs, with a learning rate of $5 \times 10^{-5}$ and a batch size of 8. The training process consisted of 50,000 steps for the domain adaptor and an additional 36,000 steps for the object injector. During inference, we set the DDIM steps as 50 and the CFG as 12.5. 

\subsection{Baselines}

We compare 3DTrajMaster with existing SOTA methods that are capable of customizing object motions: MotionCtrl~\citep{wang2024motionctrl}, Direct-a-Video~\citep{yang2024direct} and Tora~\citep{zhang2024tora}. We configure these baseline models using their best performance settings, based on their official open-sourced codebases.

\subsection{Evaluation Metric}
1) \textit{Trajectory accuracy}: Due to the absence of a pose estimator for open-world 4D objects, we limit our evaluation to only human objectives. Specifically, we utilize GVHMR~\citep{shen2024gvhmr} to estimate human poses $\{(\mathbf{R}^{est}_n,\mathbf{T}^{est}_n)\}_{n=1}^F$ and compare them with the input pose sequences $\{(\mathbf{R}^{gt}_n,\mathbf{T}^{gt}_n)\}_{n=1}^F$. We align the two trajectories at the first frame location. We follow CameraCtrl~\citep{he2024cameractrl} to estimate the rotation angle error \textbf{RotErr} and translation scale error \textbf{TransErr}, but take the average rather than the sum.
2) \textit{Video quality}: We leverage standard metrics such as Frechét Video Distance (\textbf{FVD})~\citep{unterthiner2018towards}, Frechét Image Distance (\textbf{FID})~\citep{maximilian2020fid}, and CLIP Similarity (\textbf{CLIPSIM})~\citep{wu2021godiva} to assess the video appearance.

\subsection{Evaluation Dataset}
1) \textit{Pose Sequence}: We collect 44 novel pose templates, each comprising one or more object motions.
2) \textit{Entity Description}: we use GPT to generate 20 novel human, 52 novel non-human descriptions, and 32 novel locations (refer to~\secref{subsec:evaluateprompt}), which are randomly assigned to poses to form 100 pairs (12 single-entity, 72 two-entity, and 16 three-entity each pair has one human entity).

\subsection{Comparison}

\begin{figure*}[!h]
\centering
\includegraphics[width=\linewidth]{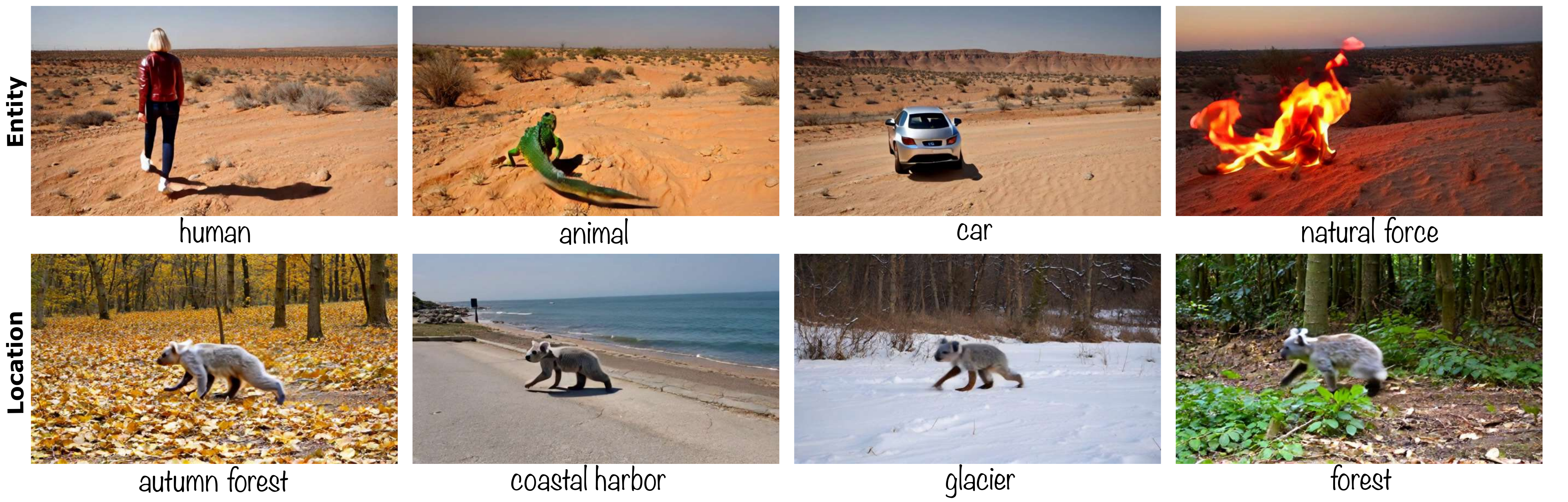}
\vspace{-2em}
\caption{\textbf{Diversity on Entity and Background.} 3DTrajMaster can control versatile entities (human, animal, car, robot, and even abstract natural force), while also generating diverse locations.}
\label{fig:diverse_entity_bg}
\end{figure*}

\begin{figure*}[!h]
\centering
\includegraphics[width=\linewidth]{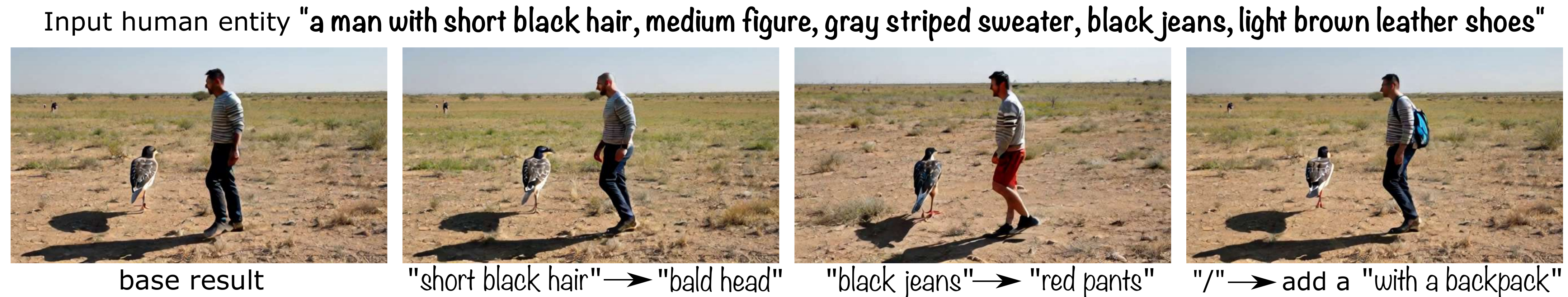}
\vspace{-2em}
\caption{\textbf{Fine-grained Editing on Human Entity Input.} 3DTrajMaster supports modifications in attributes such as hair, clothing, figure size, and so on. (Please check more in~\figref{fig:supp_diffhuman})}
\label{fig:diffhuman}
\end{figure*}

\begin{figure*}[!ht]
\centering
\includegraphics[width=\linewidth]{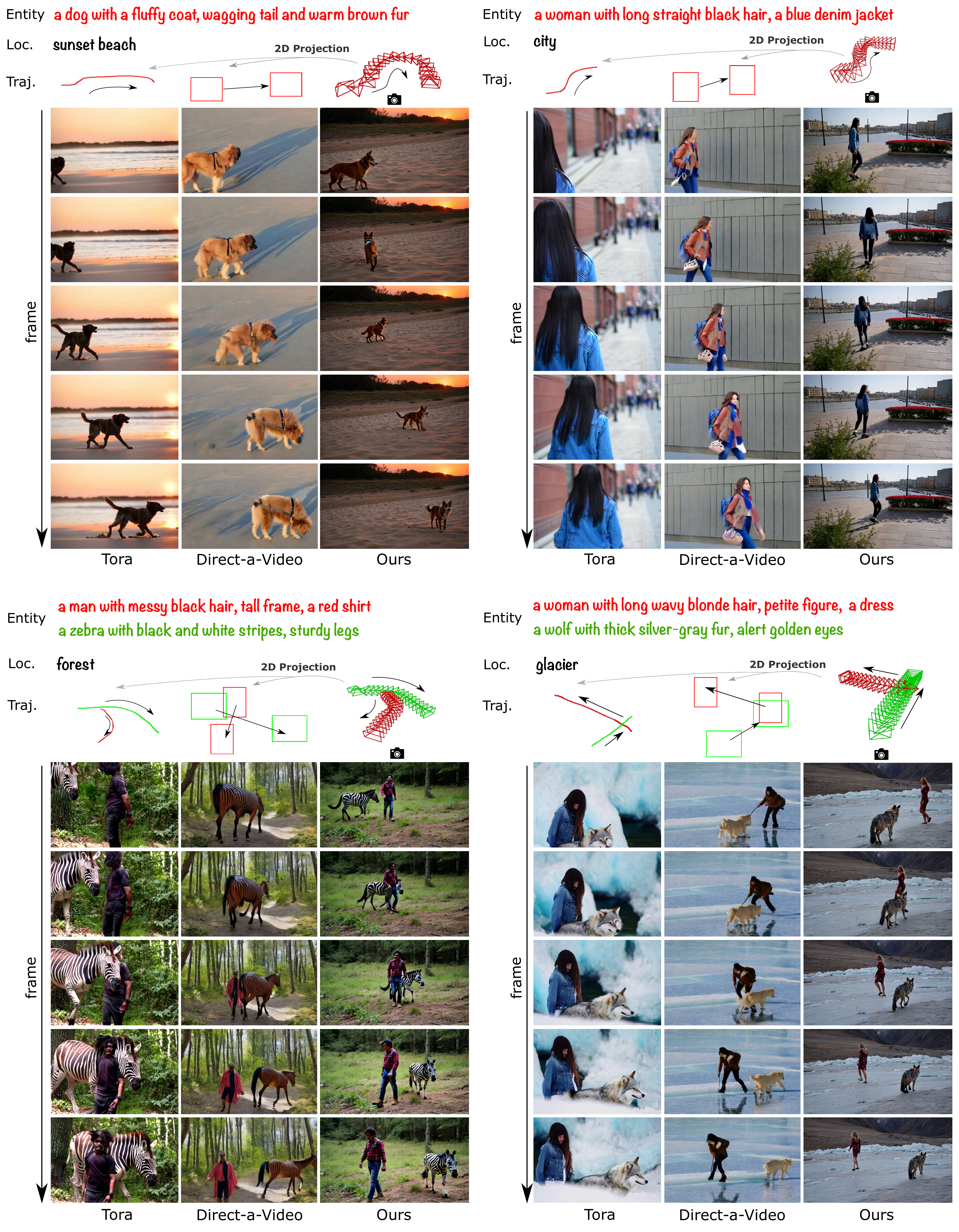}
\caption{\textbf{Qualitative Comparison on Single/Multiple Entity Motion.} 3DTrajMaster outperforms all 2D baselines by modeling 6 DoF entity motion, which can better express the inherent 3D nature of motion. In the last figure, Tora mistakenly regards the background entity as the girl entity.}
\label{fig:main_comparison}
\vspace{-2em}
\end{figure*}

\noindent \textbf{Granularity Level.} As shown in~\tabref{tab:control_level}, 3DTrajMaster can customize object location and orientation in 3D space. In contrast, 2D motion representations such as points (MotionCtrl/Tora) and bounding box (Direct-a-Video), lack awareness of the z dimension. This ambiguity becomes more problematic when handling 3D occlusion. Besides, MotionCtrl and Tora integrate multiple entities into a single 2D feature, lacking the capability to correlate individual entities with their respective trajectories (see failure case in~\figref{fig:main_comparison}). When tested on multi-entity input, Direct-a-Video (a training-free paradigm) shows particularly weak results. Furthermore, 3DTrajMaster allows for diverse entities and backgrounds (see~\figref{fig:diverse_entity_bg}), and detailed control of entity inputs (see~\figref{fig:diffhuman}).

\input{tab/control_level}

\input{tab/main_comparison}

\noindent \textbf{Quantative \& Qualitative Results.} To align with the input requirement of MotionCtrl and Direct-a-Video, we project the 3D pose trajectories onto 2D space. For baselines, we simplify the entity description, such as changing ``a man with messy black hair, tall frame, a red shirt" to ``a man" or ``a man in red". Otherwise, they may fail to generate videos with detailed descriptions. As shown in~\figref{fig:main_comparison}, in single entity settings, 3DTrajMaster generates precise entity motion, such as a 180$^{\circ}$ turn-back and a continuous inward 90$^{\circ}$ turn-around. In contrast, Tora and Direct-a-Video produce simpler motions, merely shifting objects from left to right or top-right. In the multi-entity benchmark, 3DTrajMaster successfully handles 3D occlusions, such as a man walking in front of a zebra. Direct-a-Video, however, fails in overlapping regions with mixed man and zebra. We report metric results in~\tabref{tab:main_comparison}. It is not surprising that ours significantly outperforms all baselines. 

\subsection{Ablation Study}

\input{tab/ablation_main}

\begin{figure*}[!ht]
\centering
\includegraphics[width=\linewidth]{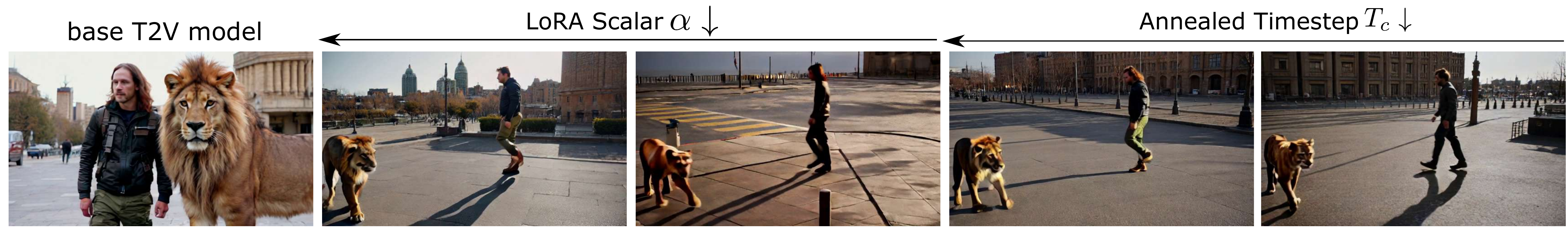}
\vspace{-1em}
\caption{\textbf{Ablation Results on Domain Adaptor (upper) and Annealed Sampling (the bottom).} We provide more experiments in~\secref{subsec:optimialhyper} to choose suitable $\alpha$ and $T_c$ to improve video quality.}
\label{fig:ablation}
\end{figure*}

\noindent \textbf{Improving Video Quality.} As illustrated in~\figref{fig:ablation} and~\tabref{tab:ablation_main}, without the video domain adaptor, the video quality deteriorates significantly, reverting to a purely UE-style appearance similar to the training set. Likewise, omitting the annealed sampling strategy results in a decline in video quality (see the beard of the lion and overall scene style). While the rotation accuracy drops slightly (0.277$\rightarrow$0.265), this is acceptable since there exist errors in evaluating open-world human poses.

\noindent \textbf{Motion Fusion Design.} As shown in~\tabref{tab:ablation_main}, replacing gated self-attention with cross-attention fusion (w/ Cross-Attn. Fusion, here we use the entity-motion bonded feature $\mathbf{Z}^\mathbf{Pe}$ as the query) or placing the object injector after the 3D self-attention layer (w/ 3D Self-Attn.) results in a slight decline in both video quality and pose sequence accuracy.

%% file: tab/control_level.tex
\begin{table}[!ht]

\vspace{-1em}
\caption{\textbf{Fine Control Comparison with Multi-Entity Input.}}
\label{tab:control_level}

\centering
\begin{threeparttable}

\resizebox{.9\textwidth}{!} 
{
\begin{tabular}{ccccccccccc}
\toprule
&  Location  & Orientation & Entity-Traj. Corresp. & Learning-based? \\

\midrule
Direct-a-Video &  \checkmark(2D) & \ding{55} & \checkmark & \ding{55} \\
MotionCtrl/Tora &  \checkmark(2D) & \ding{55} & \ding{55}  & \checkmark (not decoupled)  \\
\midrule
3DTrajMaster (Ours) & \textbf{\checkmark} (3D) & \checkmark & \checkmark & \checkmark (decoupled) \\

\bottomrule

\end{tabular}
}

\vspace{-1em}
\end{threeparttable}
\end{table}

%% file: tab/main_comparison.tex
\begin{table}[!ht]

\vspace{-2em}
\caption{\textbf{Quantative Comparison on Single/Multiple Entity Motion.} 3DTrajMaster performs better on multiple entity input since the single entity trajectory is more complex.}
\label{tab:main_comparison}

\centering
\begin{threeparttable}
\resizebox{.99\textwidth}{!} 
{
\begin{tabular}{ccccccccccc}
\toprule
& \multicolumn{2}{c}{ Single Entity } & \multicolumn{2}{c}{Multiple Entities} & \multicolumn{2}{c}{All Entities}\\
\cmidrule(lr){2-3} 
\cmidrule(lr){4-5} 
\cmidrule(lr){6-7} 
Methods &  TransErr (m) $\downarrow$ & RotErr (deg) $\downarrow$  & TransErr (m) $\downarrow$ & RotErr (deg) $\downarrow$  & TransErr (m) $\downarrow$ & RotErr (deg) $\downarrow$ \\

\midrule

Base T2V & 1.946 & 1.799 &  1.586 & 1.208 & 1.629 & 1.279 \\
MotionCtrl & 1.752 & 2.134 & 1.682 & 1.613 & 1.690 & 1.675 \\
Tora & 1.707 & 1.158 & 1.867 & 1.514 & 1.848 & 1.471 \\
Direct-a-Video & 1.632 & 1.902 & 1.391 & 0.942 & 1.420 & 1.057 \\
\midrule
3DTrajMaster & \textbf{0.456} & \textbf{0.319} & \textbf{0.390} & \textbf{0.272} & \textbf{0.398} & \textbf{0.277} \\

\bottomrule

\end{tabular}
}

\end{threeparttable}
\end{table}

%% file: tab/ablation_main.tex
\begin{table}[!ht]

\vspace{-1em}
\caption{\textbf{Ablation Study on Full Testest and Base T2V Videos (As Reference Video).} } 
\label{tab:ablation_main}

\centering
\begin{threeparttable}
\resizebox{.88\textwidth}{!} 
{
\begin{tabular}{ccccccccccc}
\toprule
&  \multicolumn{3}{c}{ Video Quality } & \multicolumn{2}{c}{ 3D Trajectory Accuracy } \\
\cmidrule(lr){2-4} 
\cmidrule(lr){5-6} 
Ablation Setting & FVD $\downarrow$ & FID $\downarrow$ & CLIPSIM $\uparrow$  &  TransErr (m) $\downarrow$ & RotErr (deg) $\downarrow$  \\

\midrule

w/ Cross-Attn. Fusion & 1673.24 & 102.13 & 32.87 & 0.453 & 0.341 \\
w/ 3D Self-Attn. & \underline{1597.51} & \underline{98.74} & \underline{33.15} & 0.427 & 0.296 \\
w/o Domain Adaptor & 2379.89 & 157.51 & 30.50 & 0.415 & 0.301 \\
w/o Annealed Sampl. & 1841.64 & 112.57 & 32.26 & \underline{0.407} & \textbf{0.265} \\
\midrule
Full Model & \textbf{1546.15} & \textbf{96.75} & \textbf{33.77} & \textbf{0.398} & \underline{0.277} \\

\bottomrule
\end{tabular}
}
\vspace{-1em}
\end{threeparttable}
\end{table}

%% file: main/conclusion.tex
\section{Conclusion}

In this work, we introduce 3DTrajMaster, a unified framework for controlling multi-entity motions in 3D space, with motion representation as 6DoF location and rotation sequences. Our flexible object injector establishes entity-wise correspondence and allows flexible editing of entity descriptions.

\noindent \textbf{Limitation.} 
Generalizable entities, like animals, cannot be edited with the same level of granularity as humans. This limitation can be addressed by constructing more diverse and detailed 3D assets of the same category. Currently, the model is constrained to global motion patterns; however, fine-grained local motions (\eg, human dancing or waving hands) and interactions between different entities (\eg, a man picking up a dog) can also be modeled similarly to our 6 DoF motions with structured motion patterns. 
At present, our model can only generate limited entities ($\leq$3) at a time, but this can be improved with more powerful video foundation models and paired datasets.

%% file: main/acknowledgment.tex
\section*{Acknowledgments}

We thank Jinwen Cao, Yisong Guo, Haowen Ji, Jichao Wang, and Yi Wang from Kuaishou Technology for their help in constructing our 360$^{\circ}$-Motion Dataset. As for the fruitful discussion, we thank Yuzhou Huang, Qinghe Wang, Runsen Xu, Zeqi Xiao, and Zhouxia Wang.

%% file: supp/supp.tex
\renewcommand{\thetable}{R\arabic{table}}
\renewcommand\thefigure{S\arabic{figure}}

\appendix
\section*{Appendix}

\section{Internal Video Diffusion Model for Research Purpose} \label{subsec:videobackbone}

\begin{figure*}[ht]
\centering
\includegraphics[width=.9\linewidth]{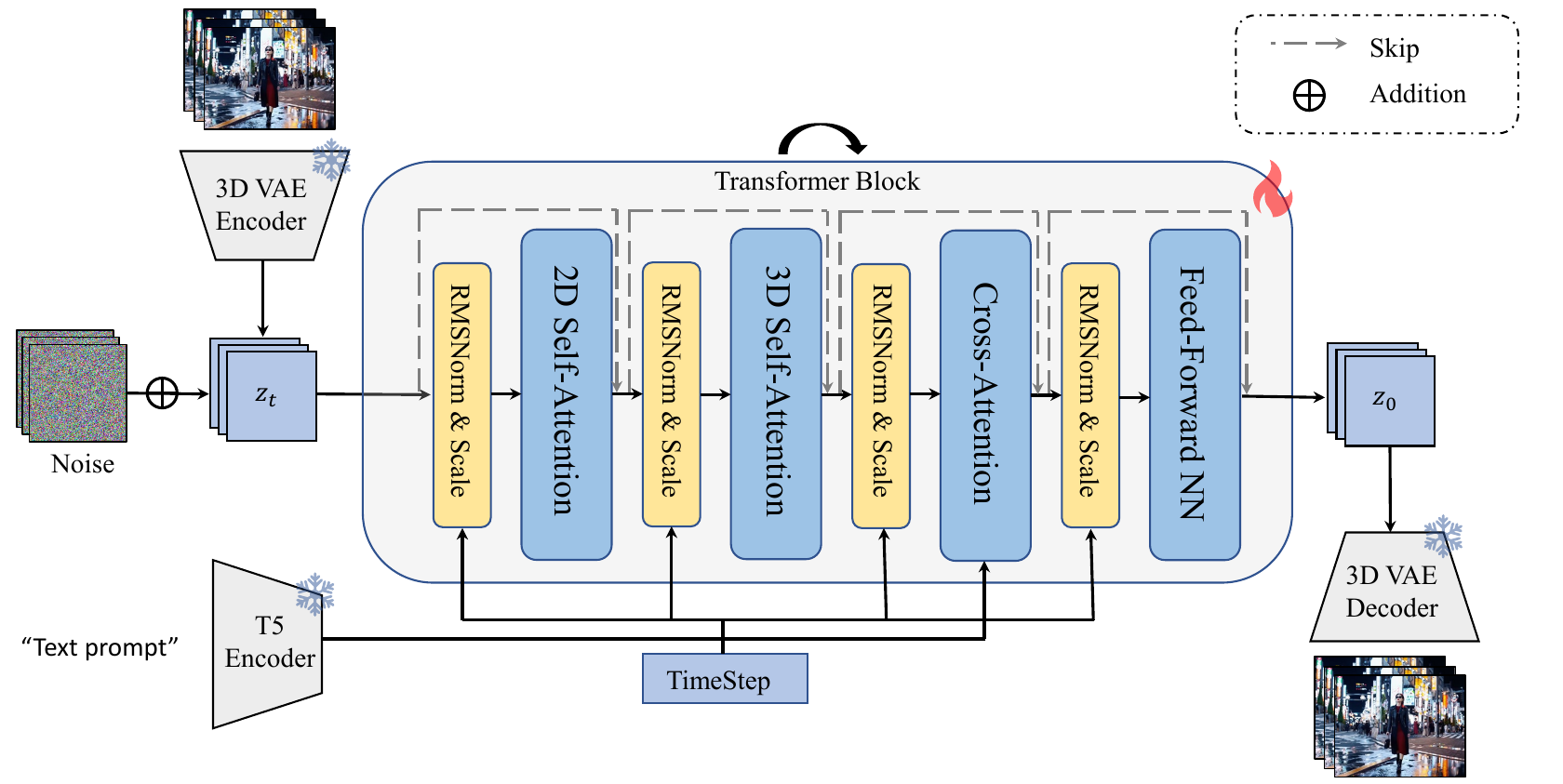}
\caption{\textbf{Our Video Latent Diffusion Model Backbone}}
\label{fig:dit_backbone}
\end{figure*}

Our model is a transformer-based latent diffusion model, as illustrated in the~\figref{fig:dit_backbone}. Initially, we employ a 3D VAE to transform videos from the pixel level to a latent space, upon which we construct a transformer-based video diffusion model~\citep{peebles2023scalable}. Previous models, which rely on UNets~\citep{blattmann2023stable,chen2023videocrafter1,guo2023animatediff} or transformers~\citep{ma2024latte}, typically incorporate an additional 1D temporal attention module for video generation, and such spatial-temporally separated designs do not yield optimal results. Instead, we replace the 1D temporal attention with 3D self-attention~\citep{gupta2023photorealistic}, enabling the model to more effectively perceive and process spatiotemporal tokens, thereby achieving a high-quality and coherent video generation model. Specifically, we map the timestep to a scale, thereby applying RMSNorm to the spatiotemporal tokens before each attention or feed-forward network (FFN) module.

\section{Additional Related Work}

\noindent \textbf{Injecting Control into Video Foundation Models.} \textit{(1) Learning-based}: The control signals are typically projected into latent embeddings via an extra encoder (\eg, learnable convolutional/linear/attention/LoRA layers, or frozen pre-trained feature encoder), which are then integrated into the base model architecture through concatenation, addition, or insertion. VideoComposer[1] employs a unified STC-encoder and CLIP model to feed multi-modal input conditions (textual, spatial, and temporal) into the base T2V model. MotionCtrl~\citep{wang2024motionctrl} introduces camera motion by fine-tuning specific layers of the base U-Net, and object motion via additional convolutional layers. CameraCtrl~\citep{he2024cameractrl} enhances this approach by incorporating ControlNet~\citep{zhang2023adding}'s philosophy, using an attention-based pose encoder to fuse camera signals in the form of Plücker embeddings while keeping the base model frozen. Similarly, SparseCtrl~\citep{guo2023sparsectrl} learns an add-on encoder to integrate control signals (RGB, sketch, depth) into the base model. Tora~\citep{zhang2024tora} employs a trajectory encoder and plug-and-play motion fuser to merge 2D trajectories with the base video model. MotionDirector[7] leverages spatial and temporal LoRA layers to learn desired motion patterns from reference videos.
\textit{(2) Training-free}: These methods modify attention layers or video latents to adjust control signals in a computationally efficient manner. However, training-free methods often suffer from poor generalization and require extensive trial-and-error. Direct-a-video~\citep{yang2024direct} amplifies or suppresses attention in spatial cross-attention layers to inject box guidance, while FreeTraj~\citep{qiu2024freetraj} embeds target trajectories into the low-frequency components and redesigns reweighting strategies across attention layers. MOFT~\citep{xiao2024video} extracts motion priors by removing content correlation and applying motion channel filtering, and then alters the sampling process using the reference MOFT. 

\section{Additional Applications}

We outline our potential applications in various areas as follows.

1) \textbf{Film:} Reproduce the character's classic moves. We can extract the human poses from a given video and apply them to different entities and backgrounds using the capabilities of our model.

2) \textbf{Autonomous Driving:}  Simulate dangerous safety accidents, such as two cars colliding and a car hitting a person.

3) \textbf{Embodied AI:} Generate a vast number of videos with diverse entity and trajectory inputs to train a general 4D pose estimator, especially for non-rigid objects.

4) \textbf{Game:} Train a character ID, such as Black Myth Wukong, through LoRA, and then drive the character movement with different trajectories.

\section{Clarification of the Limited Entity Number ($\leq$3)} 

Currently, our method is limited to generating up to 3 entities, as outlined in the `Limitation' section of the paper. This constraint is primarily due to the capabilities of the video foundation model rather than the training data. While it is relatively easy to generate $\gg$2 entities of the same category (e.g., ``a group of people/cars/animals") in the video, it becomes much more challenging to generate $\gg$2 entities, each differing greatly from the others, through the text input as T5 text encoder tends to mix the textual features of different entities. Thus it becomes hard to associate specific trajectories with their corresponding text entities. Based on empirical studies with video foundation models, we chose to limit the number of entities to 3 in our work. Regarding data construction, it is easy to include more entities with their paired trajectories in our procedure UE platform pipeline. However, the key limitation is that the video foundation model struggles to generate such a diverse set of entities simultaneously. Furthermore, many prior works, such as Tora, MotionCtrl, and Direct-a-video also focus on a limited number of entities.

\section{Dataset Illustration} 

\subsection{360$^{\circ}$-Motion Dataset Data.} \label{subsec:360motiondataset}
\begin{figure*}[!ht]
\centering
\includegraphics[width=\linewidth]{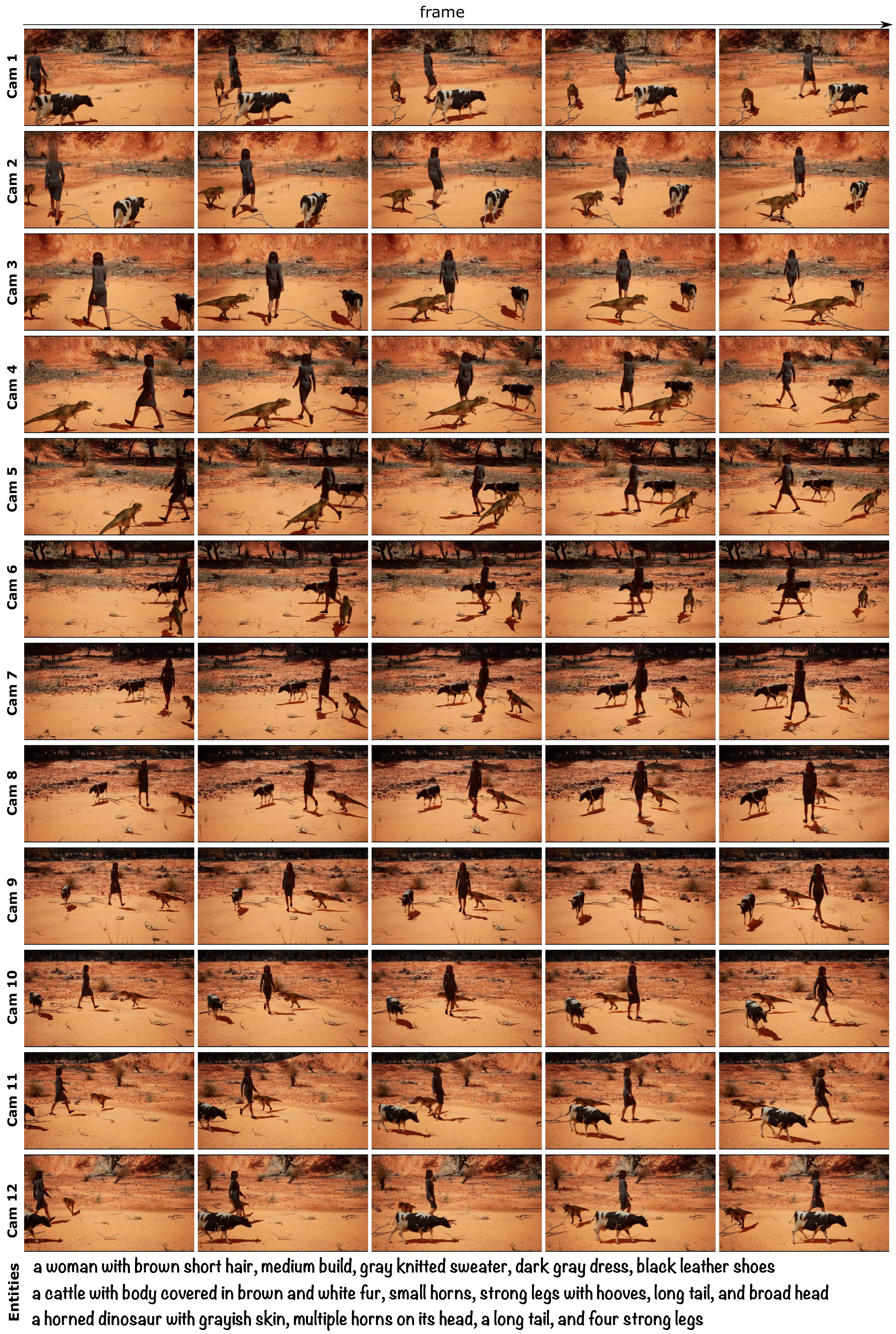}
\caption{\textbf{A Sample from our \textit{360$^{\circ}$-Motion} Captured with 12 Evenly-Surrounded Cameras.}}
\label{fig:supp_dataset}
\end{figure*}

We show a sample in~\figref{fig:supp_dataset} captured with 12 evenly-surrounded cameras. Each camera shoots a clip of 100 frames at 384$\times$672 resolutions. During training, we discard the initial 10 frames to eliminate potential blurring and noise caused by 3D model initialization in the UE platform.

\subsection{Unbalanced Entity Distribution in Common Video Datasets} \label{subsec:unbalanceddataset}

\begin{figure*}[!ht]
\centering
\includegraphics[width=\linewidth]{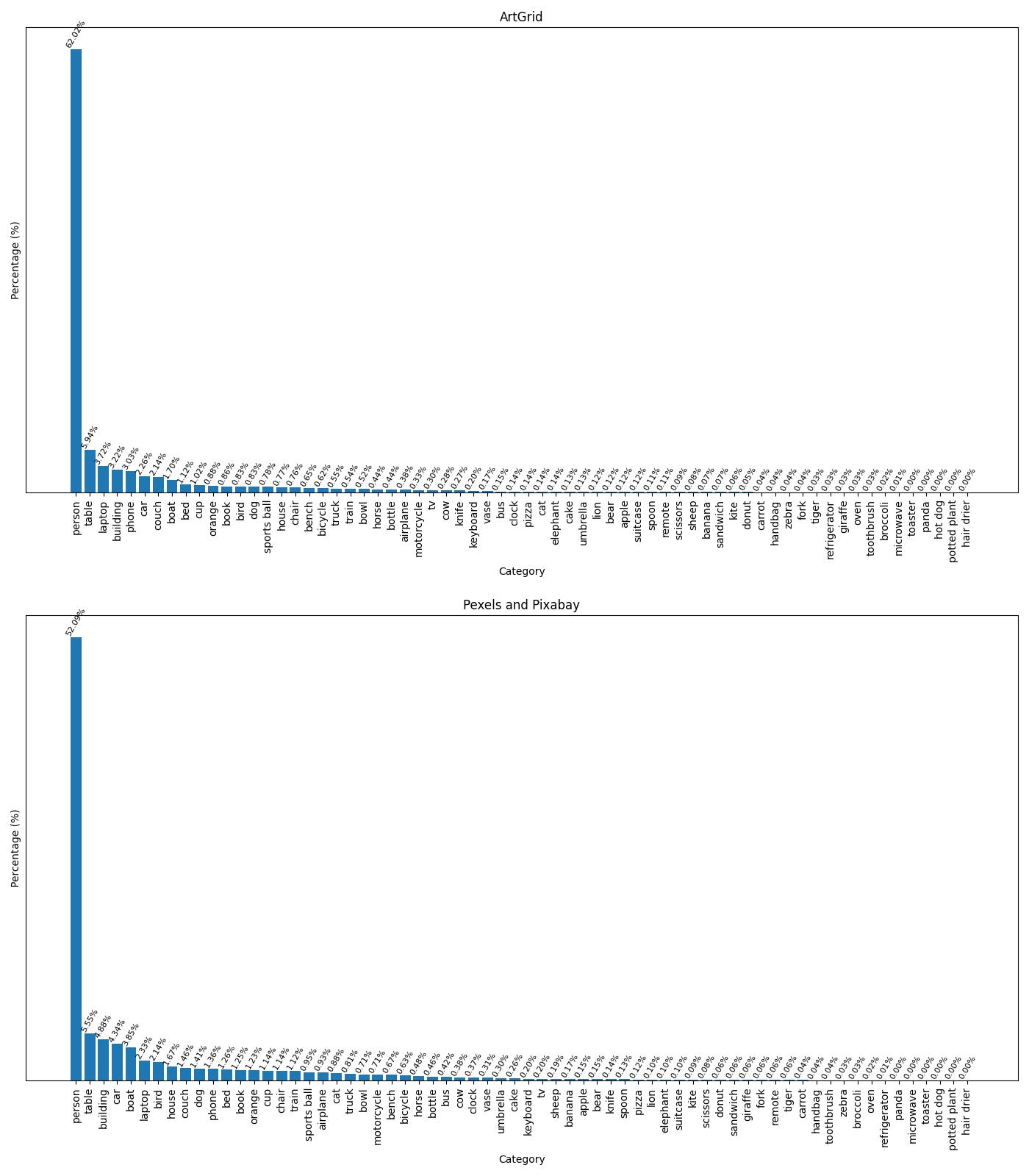}
\caption{\textbf{Entity Distribution Over 60 Classes in Artgrid, Pixabay, and Pexels.}}
\label{fig:entity_distribution}
\end{figure*}

In high-quality video datasets like Artgrid, Pixabay, and Pexels\footnote{Artgrid: https://artgrid.io/, Pixabay: https://www.videvo.net/, Pexels: https://www.pexels.com/}, the issue of category imbalance is highly pronounced and poses significant challenges. We analyze the aforementioned three datasets by first captioning the videos using QWen-VL~\citep{Qwen-VL}. Subsequently, we employ the spaCy\footnote{spaCy: https://spacy.io/} library to extract noun chunks from the video captions, which serve as entity words. We predefine over 60 classes as keywords for entity filtering. As illustrated in the~\figref{fig:entity_distribution}, certain categories (\eg, humans) constitute a disproportionately large share of the entity objects, thereby constraining the model's ability to generalize to other categories that appear less frequently.

\subsection{GPT-Generated Evaluation Prompts} \label{subsec:evaluateprompt}

The human prompts, non-human (animal, car, robot) prompts, and location prompts for evaluation are provided in~\tabref{tab:human_prompts}, ~\tabref{tab:nonhuman_prompts_1}\&\tabref{tab:nonhuman_prompts_2}, and~\tabref{tab:location_prompts} respectively.

\input{tab/supp_gptprompt_human}
\input{tab/supp_gptprompt_nonhuman_1}
\input{tab/supp_gptprompt_nonhuman_2}
\input{tab/supp_gptprompt_location}

\section{More Experiments}

\subsection{Fine-grained Entity Prompt Input}

\begin{figure*}[!ht]
\centering
\includegraphics[width=\linewidth]{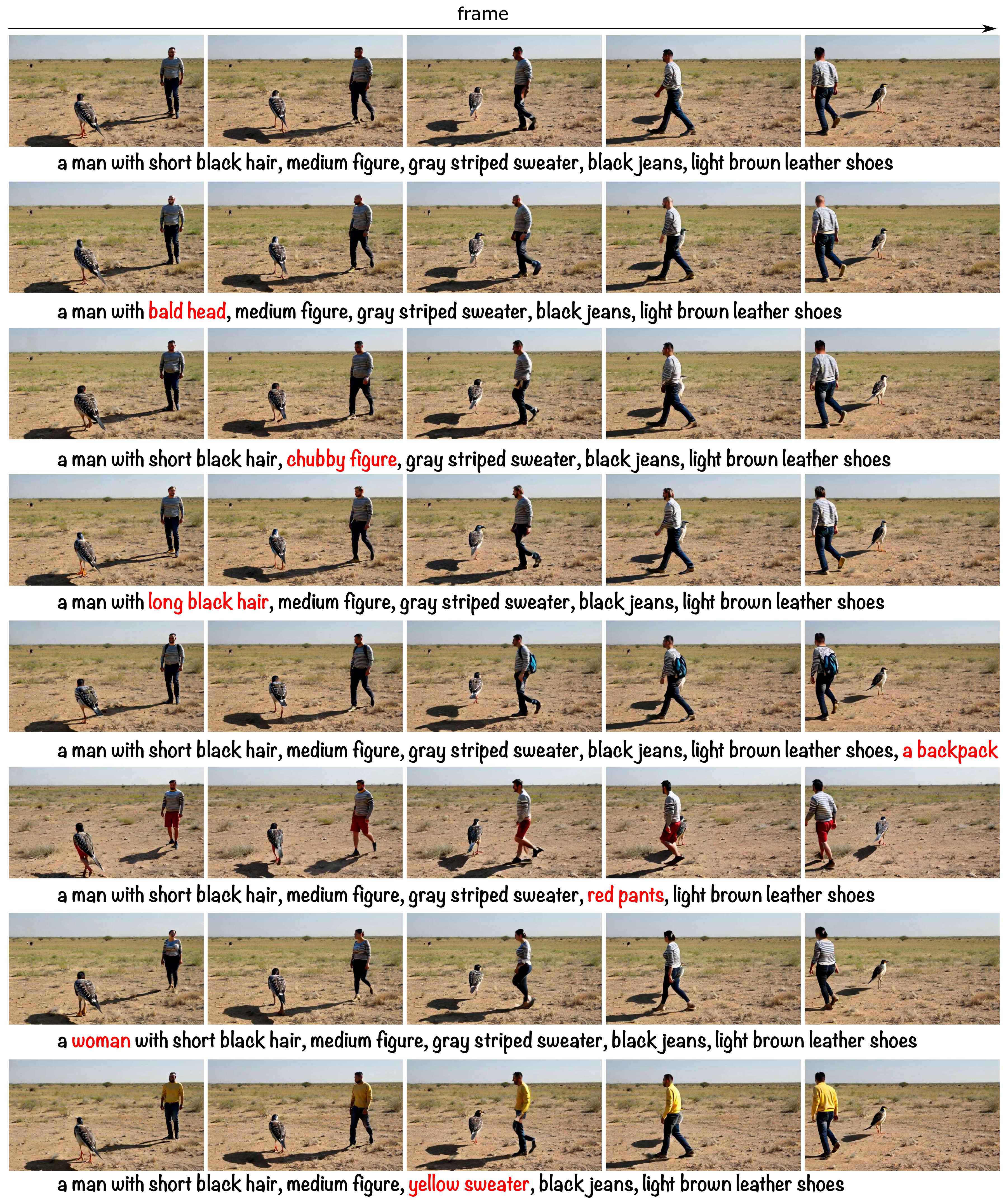}
\caption{\textbf{Flexible Entity Editing in Input Text Prompts.} The other entity, \textit{``a swift falcon with blue-gray feathers, sharp talons, and keen yellow eyes focused on its prey below"} remains fixed while varying the human entity descriptions.}
\label{fig:supp_diffhuman}
\end{figure*}

We provide additional samples in~\figref{fig:supp_diffhuman} to demonstrate that 3DTrajMaster supports fine-grained entity customization. The description of the man can be flexibly modified by adjusting attributes such as hair, gender, physique, clothing, and accessories.

\subsection{Ablation Study}

\subsubsection{Optimal Hyperparameters} \label{subsec:optimialhyper}

In the main paper, we propose a video domain adaptor and an annealed sampling strategy to mitigate video domain shifts from our constructed UE datasets. However, completely removing the LoRA adaptor (as the learned motion and domain bias are coupled to some extent) or the inserted motion guidance will result in a decline in 3D trajectory accuracy. Thus, applying video enhancement techniques with appropriate dropping is crucial. To this end, we begin with a randomly initialized parameter group: $T_{c}=10, \alpha=0.2, TS=72,000$. We perform ablation experiments on our evaluation subset. As shown in~\tabref{tab:ablation_annealed_tc},~\tabref{tab:ablation_lora_alpha}, and~\tabref{tab:ablation_ts}, the video quality exhibits a monotonically decreasing trend as these hyperparameters increase. In contrast, 3D trajectory accuracy initially drops sharply but stabilizes in the later stages. To balance the degradation of visual quality with maintaining pose accuracy, we select an optimal parameter group: $T_{c}=25, \alpha=0.4, TS=36,000$ as our default inference setting.

\input{tab/ablation_annealed_tc}
\input{tab/ablation_lora_alpha}
\input{tab/ablation_training_step}

\subsubsection{Negative Pose Condition as Static Motions} \label{subsec:negpose}

We find that setting negative pose sequences as static motions $\{(\hat{\mathbf{P}}_n)_{n=1}^N|\hat{\mathbf{P}}_n=\mathbf{P}_0,\forall n\}$ rather than positive motion sequences $\{(\mathbf{P}_n)_{n=1}^N\}$ can further improve pose accuracy, as shown in~\tabref{tab:ablation_neg_pose}. We infer that the model captures underlying 3D motion representations from the randomly generated 3D trajectories. However, we do not adopt this approach due to the decline in video quality.

\input{tab/ablation_negative_pose}

\subsubsection{Qualitative Feedback from Human Users}

We conducted a questionnaire survey and collected 53 samples to form user preference comparisons. Each participant received a reward of 0.80 USD and spent approximately 5 minutes completing the questionnaire, which assessed four dimensions: (1) video quality, (2) trajectory accuracy, (3) entity diversity, and (4) background diversity. In~\tabref{tab:user_study}, we report the proportion of users who preferred our model over the baselines.

\begin{table}[!ht]

\caption{User Preference Comparisons.}
\label{tab:user_study}

\centering
\begin{threeparttable}
\resizebox{.57\textwidth}{!} 
{
\begin{tabular}{c|c|c|c}
\toprule
Method & MotionCtrl & Direct-a-Video &  Tora \\
\midrule 3DTrajMaster & $47.2 \%$ & $56.6 \%$ & $81.1 \%$ \\
\bottomrule
\end{tabular}
}

\end{threeparttable}
\end{table}

\subsubsection{Generalizable Entity Prompts\&3D Trajectories} \label{subsec:generalizableresults}

\begin{figure*}[!ht]
\centering
\includegraphics[width=.94\linewidth]{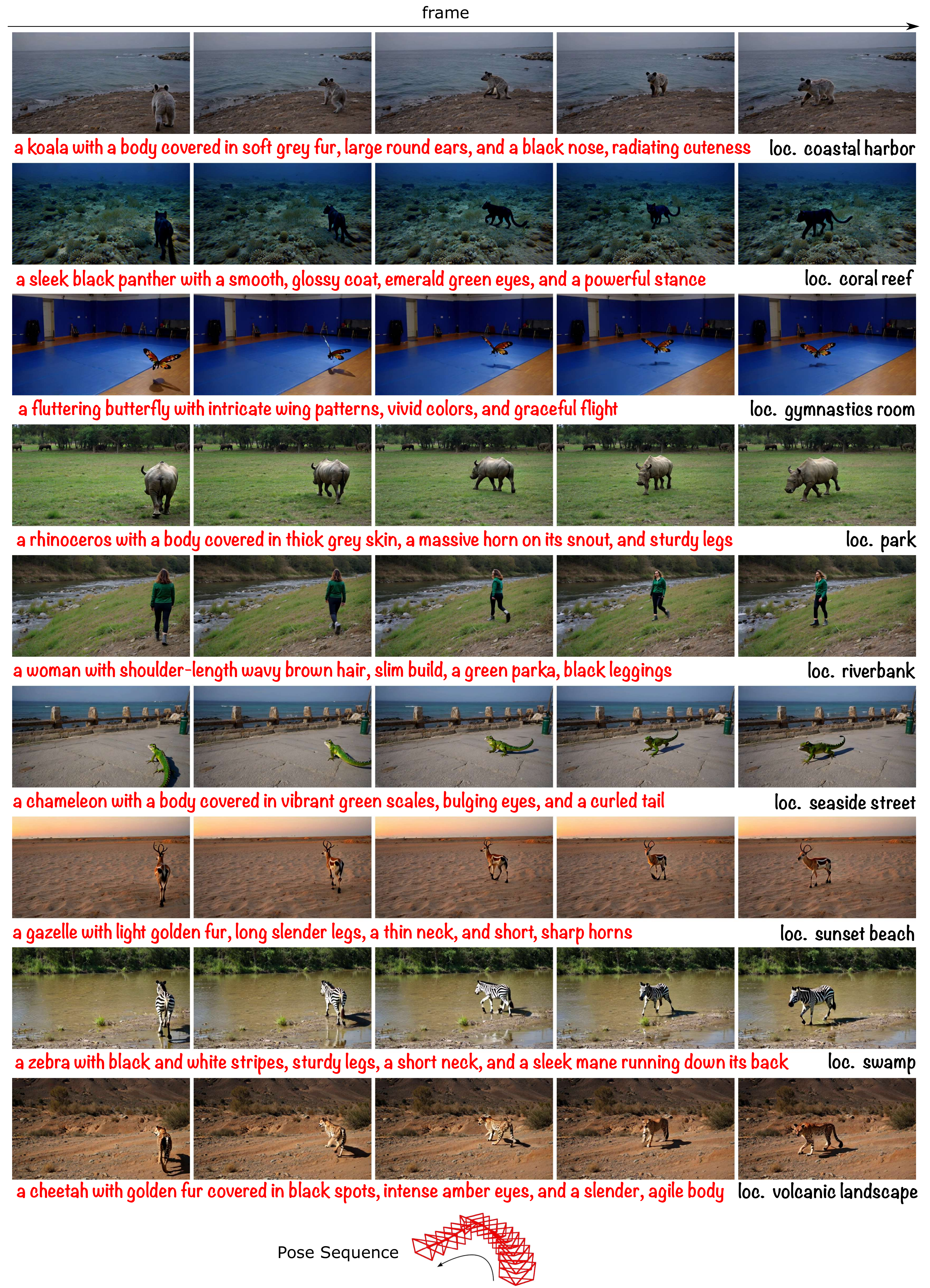}
\caption{\textbf{Generalizable Results with Novel 3D Trajectories \& Entity Prompts (1/20)}}
\label{fig:sub_generalizable_start}
\end{figure*}

\begin{figure*}[!ht]
\centering
\includegraphics[width=.93\linewidth]{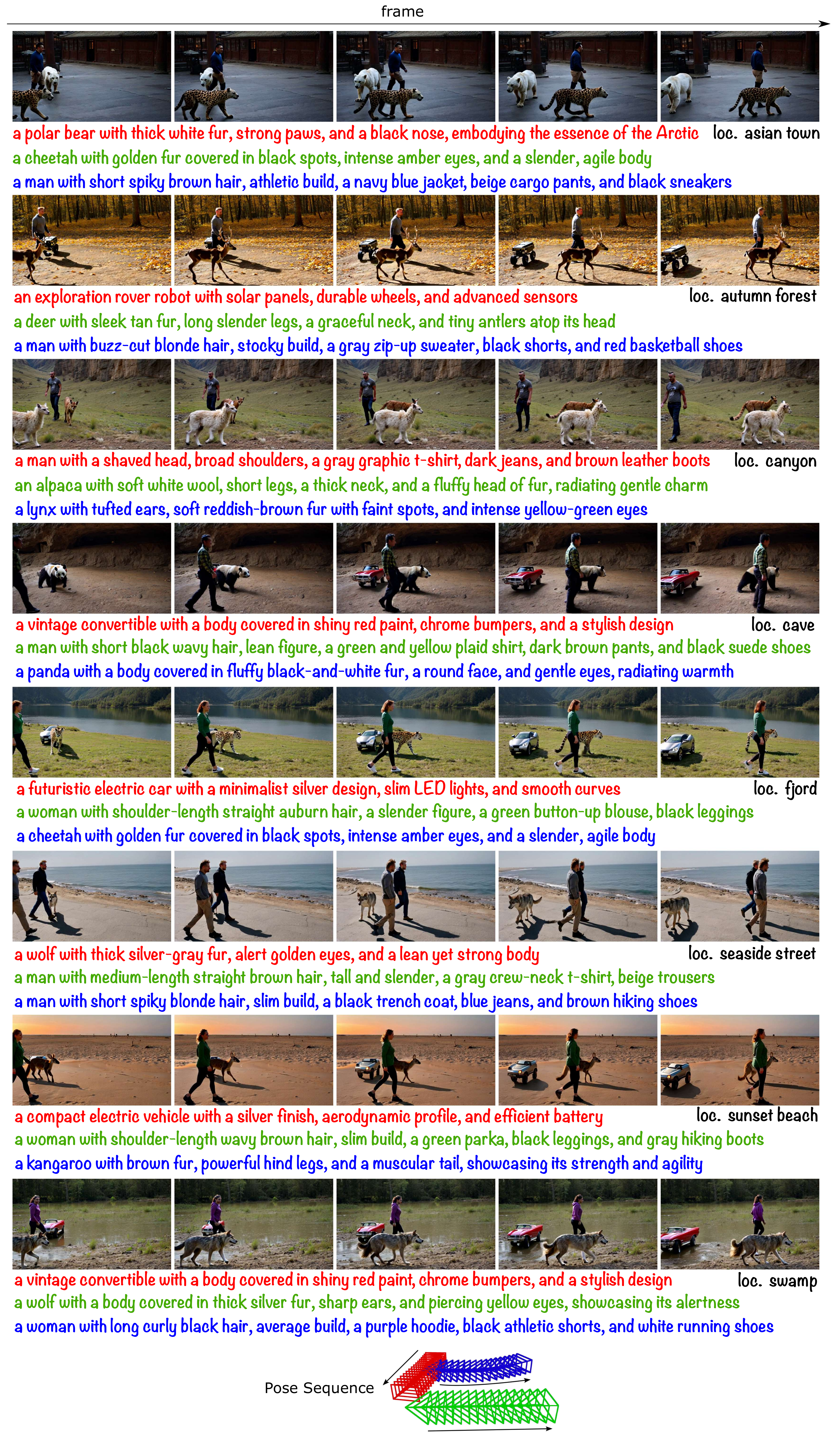}
\caption{\textbf{Generalizable Results with Novel 3D Trajectories \& Entity Prompts (20/20)}}
\label{fig:sub_generalizable_end}
\end{figure*}

We provide more generalizable results with novel entity prompts generated by GPT and 3D trajectories, as shown in~\figref{fig:sub_generalizable_start} to~\figref{fig:sub_generalizable_end}. Each text prompt consists of one to three entities. \textcolor{red}{\textit{(We kindly urge readers to check the visual results in the our website)}}.

\begin{figure*}[!ht]
\centering
\includegraphics[width=.93\linewidth]{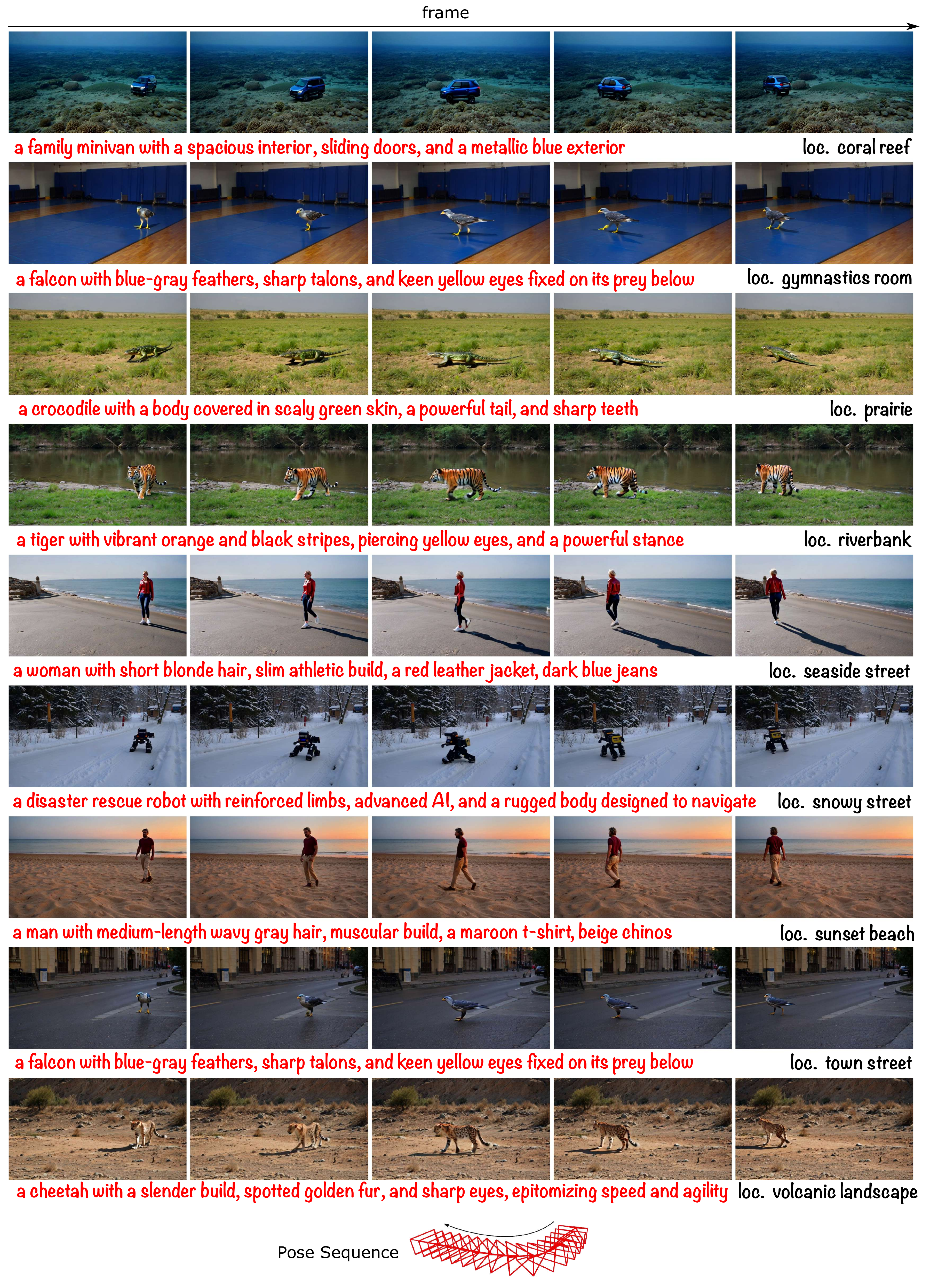}
\caption{\textbf{Generalizable Results with Novel 3D Trajectories \& Entity Prompts (2/20)}}
\end{figure*}

\begin{figure*}[!ht]
\centering
\includegraphics[width=.93\linewidth]{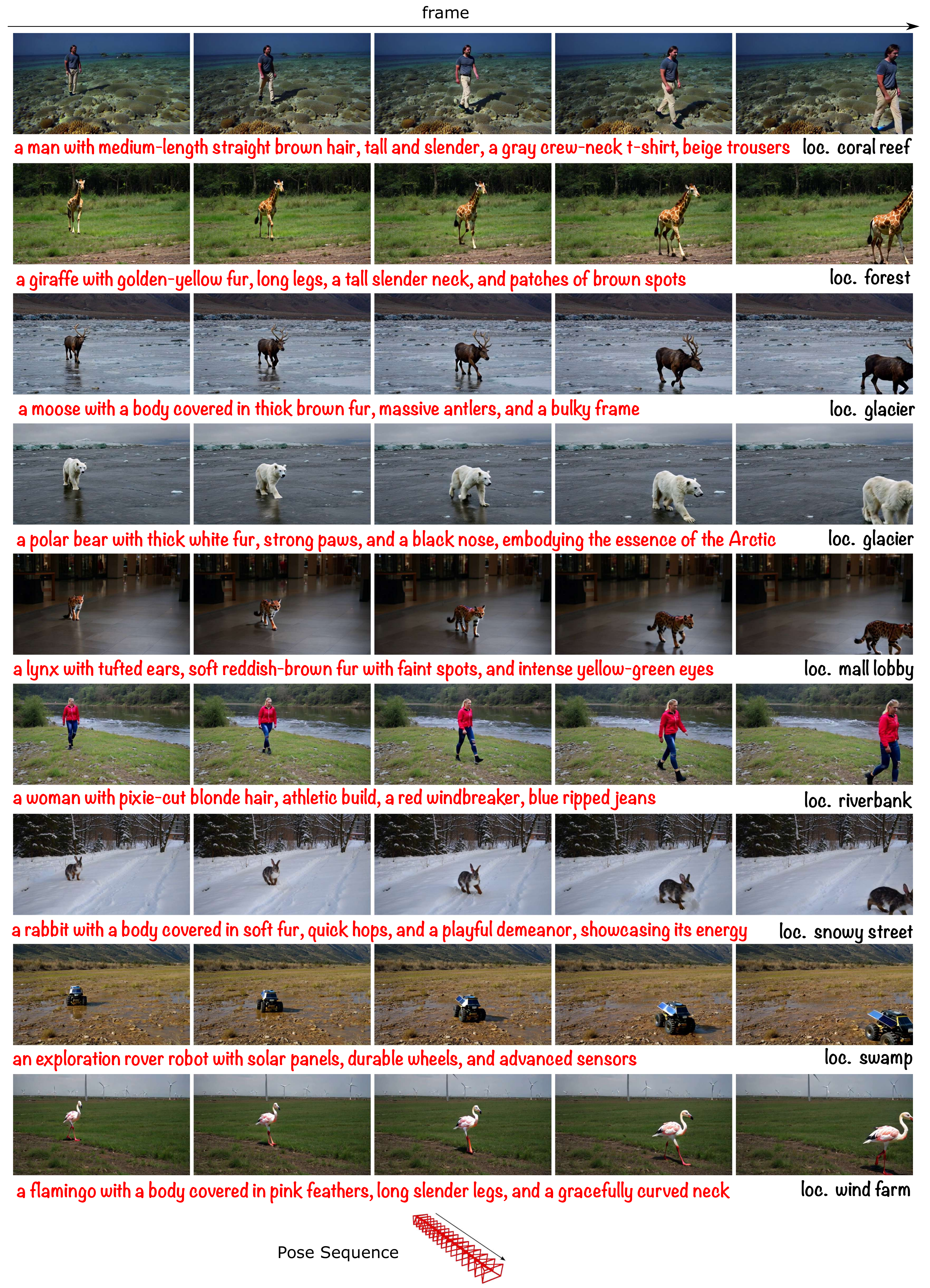}
\caption{\textbf{Generalizable Results with Novel 3D Trajectories \& Entity Prompts (3/20)}}
\end{figure*}

\begin{figure*}[!ht]
\centering
\includegraphics[width=.93\linewidth]{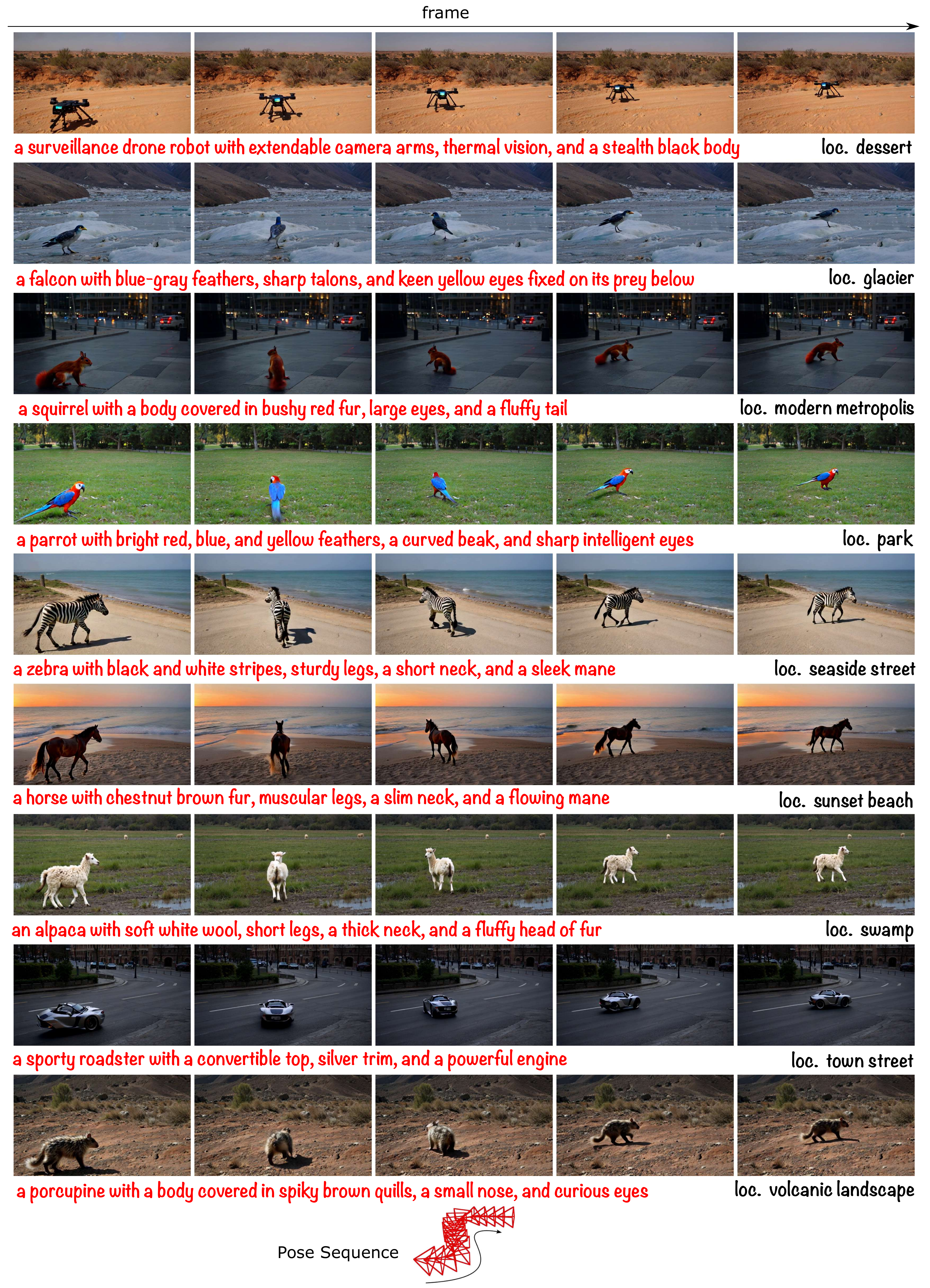}
\caption{\textbf{Generalizable Results with Novel 3D Trajectories \& Entity Prompts (4/20)}}
\end{figure*}

\begin{figure*}[!ht]
\centering
\includegraphics[width=.93\linewidth]{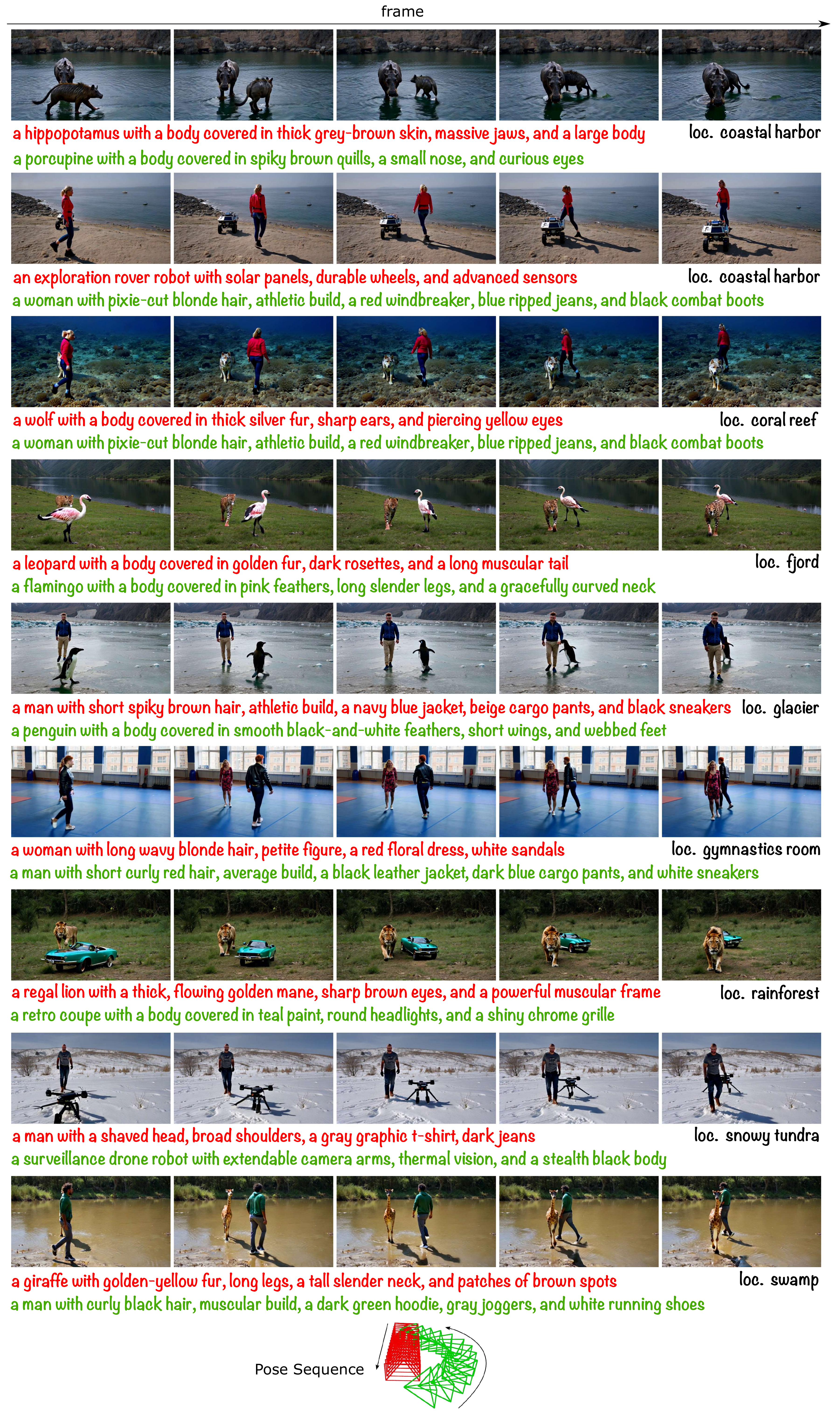}
\caption{\textbf{Generalizable Results with Novel 3D Trajectories \& Entity Prompts (5/20)}}
\end{figure*}

\begin{figure*}[!ht]
\centering
\includegraphics[width=.93\linewidth]{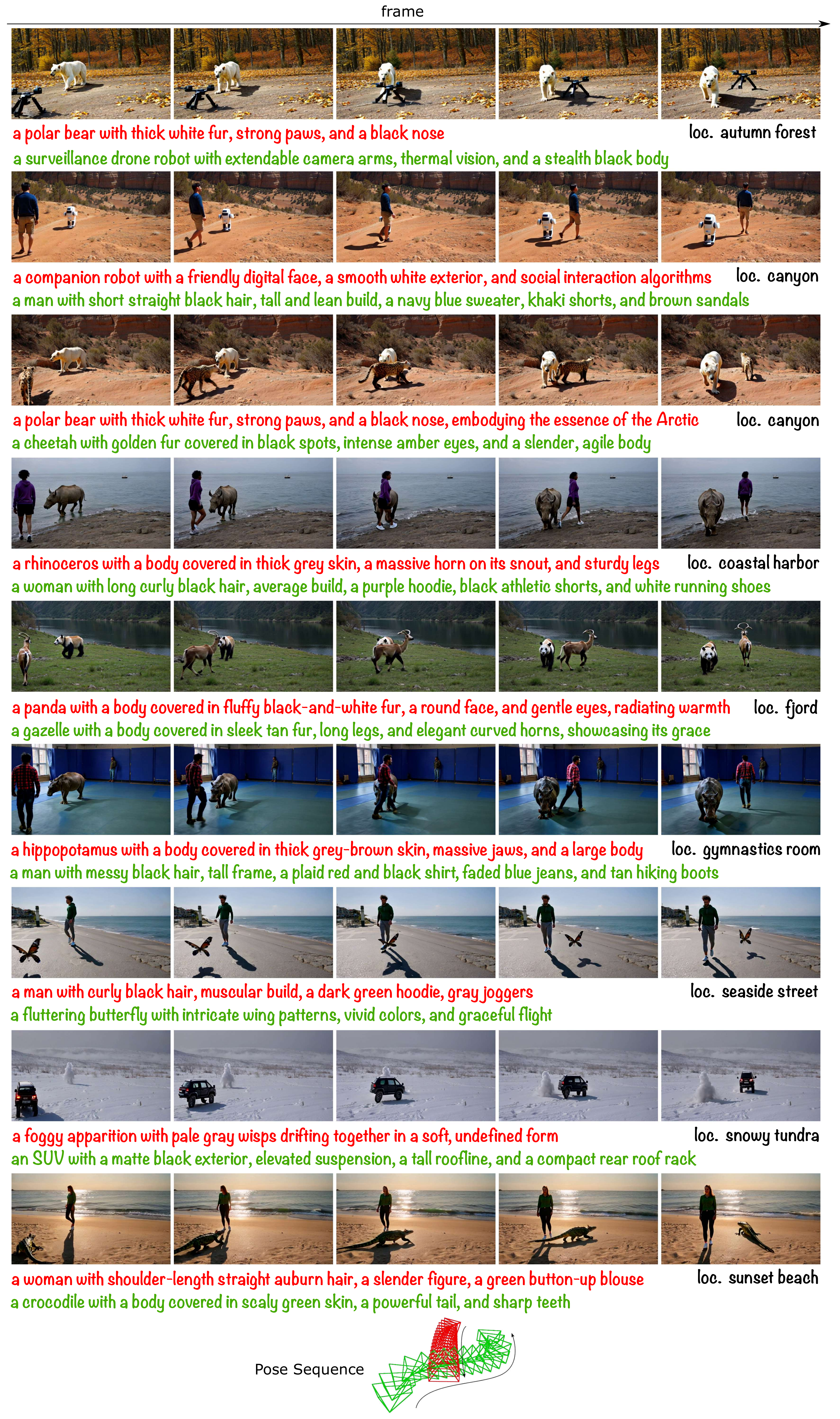}
\caption{\textbf{Generalizable Results with Novel 3D Trajectories \& Entity Prompts (6/20)}}
\end{figure*}

\begin{figure*}[!ht]
\centering
\includegraphics[width=.93\linewidth]{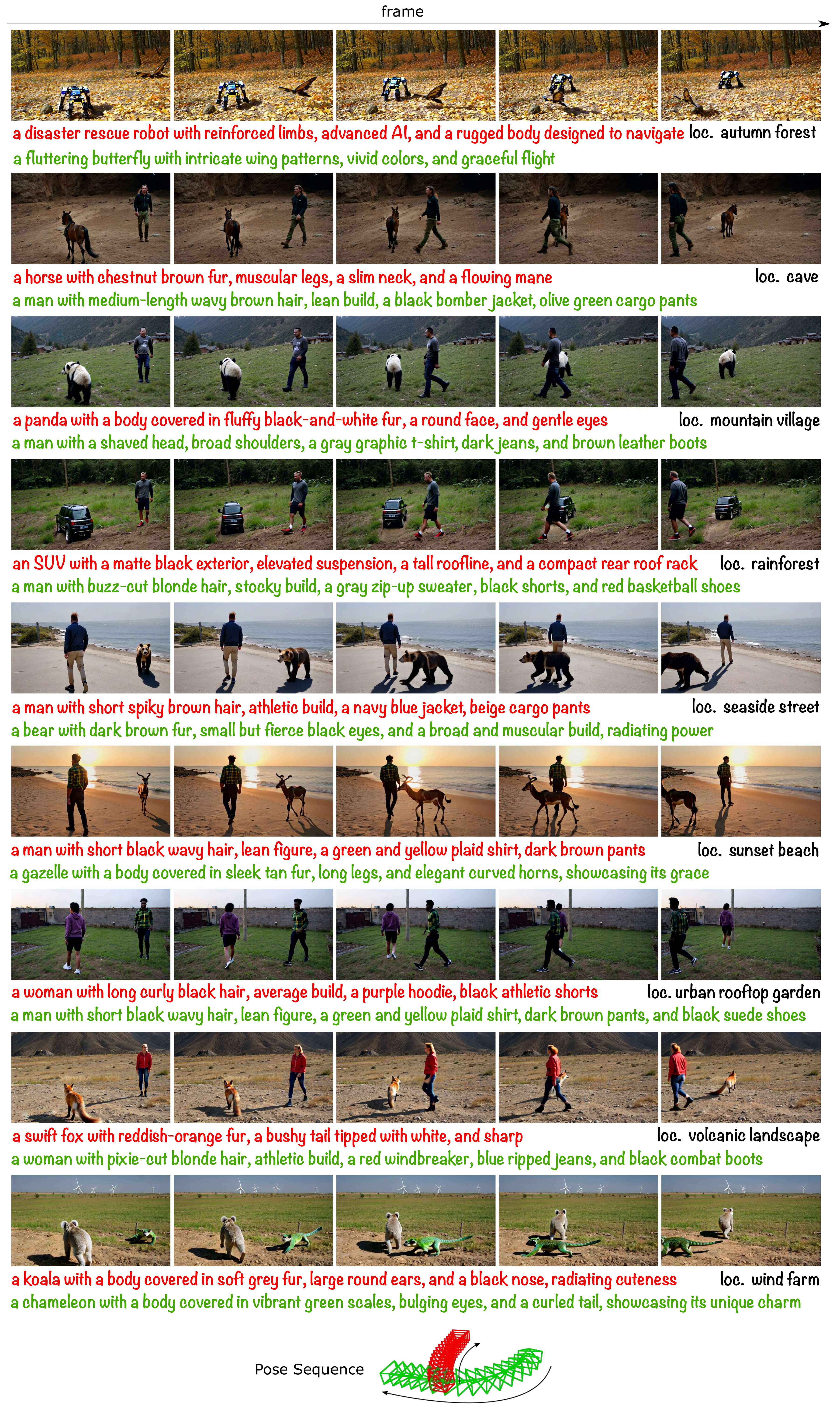}
\caption{\textbf{Generalizable Results with Novel 3D Trajectories \& Entity Prompts (7/20)}}
\end{figure*}

\begin{figure*}[!ht]
\centering
\includegraphics[width=.93\linewidth]{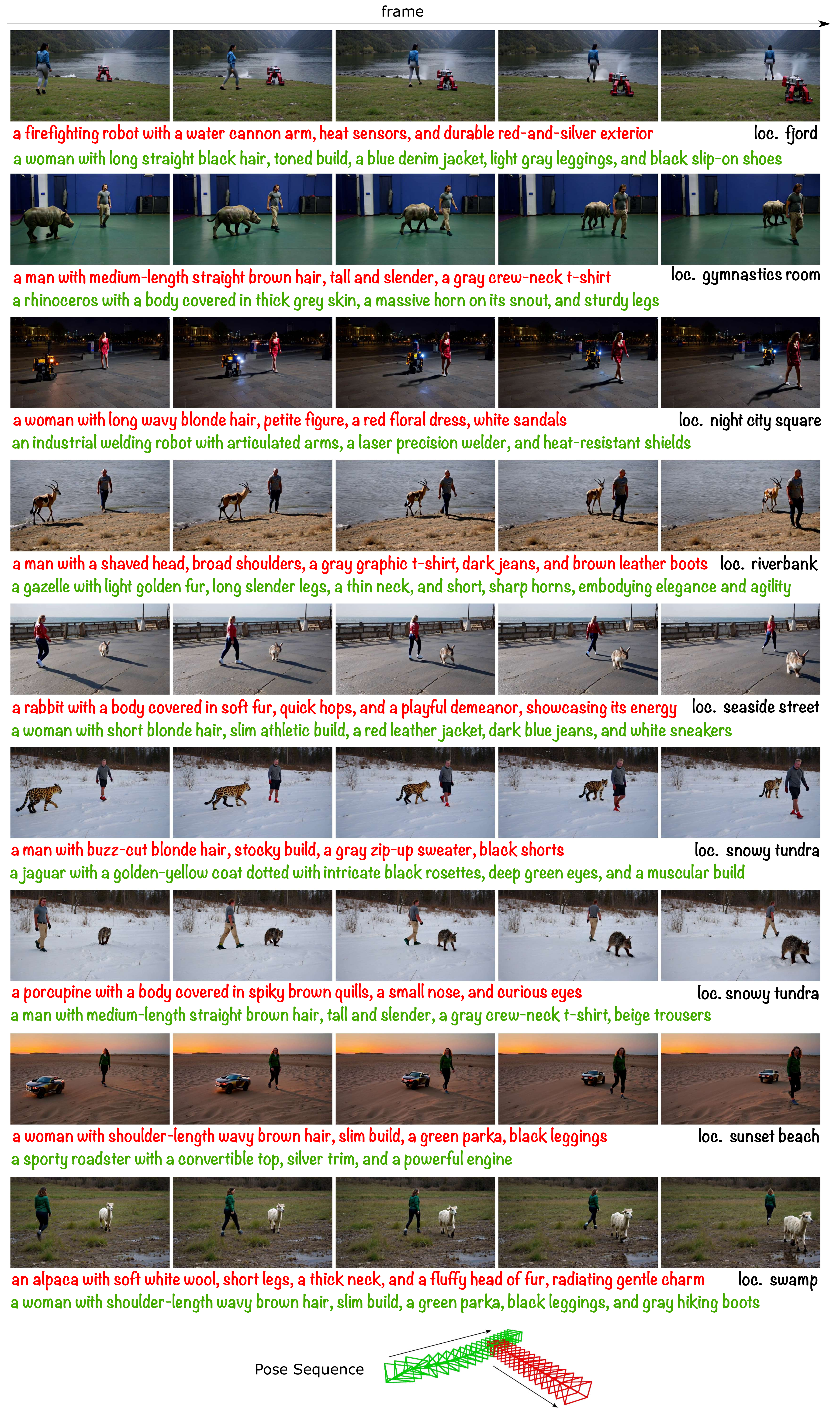}
\caption{\textbf{Generalizable Results with Novel 3D Trajectories \& Entity Prompts (8/20)}}
\end{figure*}

\begin{figure*}[!ht]
\centering
\includegraphics[width=.93\linewidth]{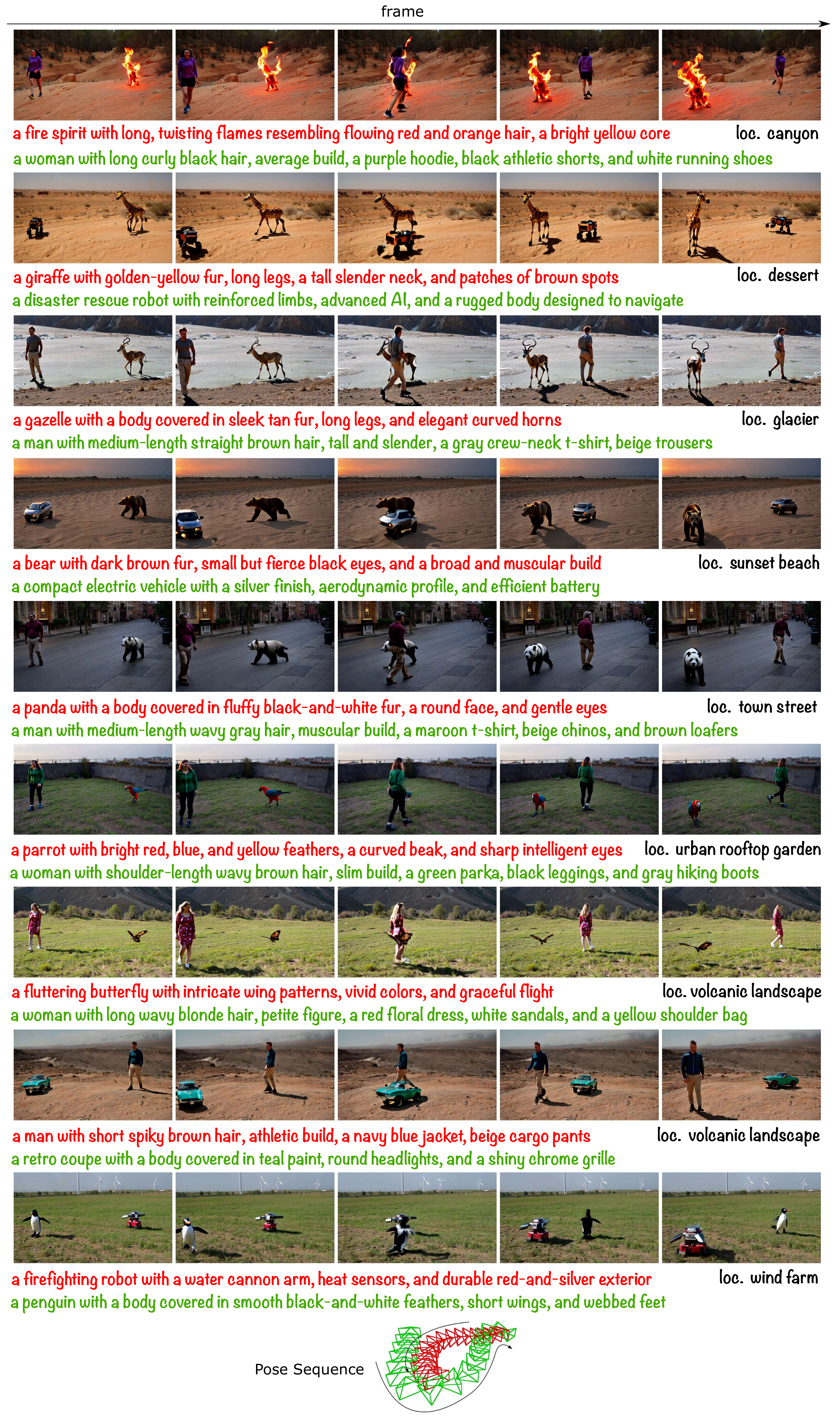}
\caption{\textbf{Generalizable Results with Novel 3D Trajectories \& Entity Prompts (9/20)}}
\end{figure*}

\begin{figure*}[!ht]
\centering
\includegraphics[width=.93\linewidth]{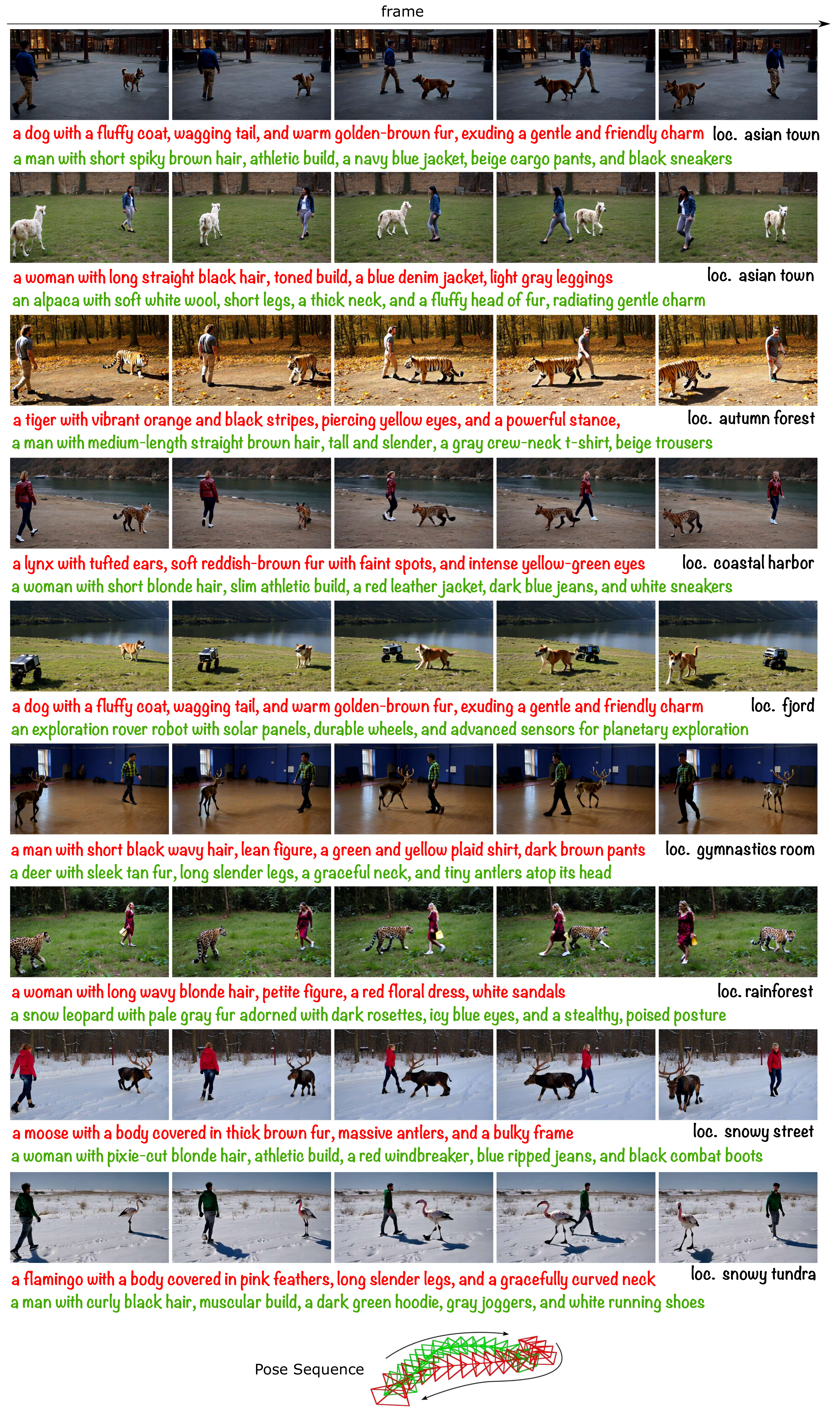}
\caption{\textbf{Generalizable Results with Novel 3D Trajectories \& Entity Prompts (10/20)}}
\end{figure*}

\begin{figure*}[!ht]
\centering
\includegraphics[width=.93\linewidth]{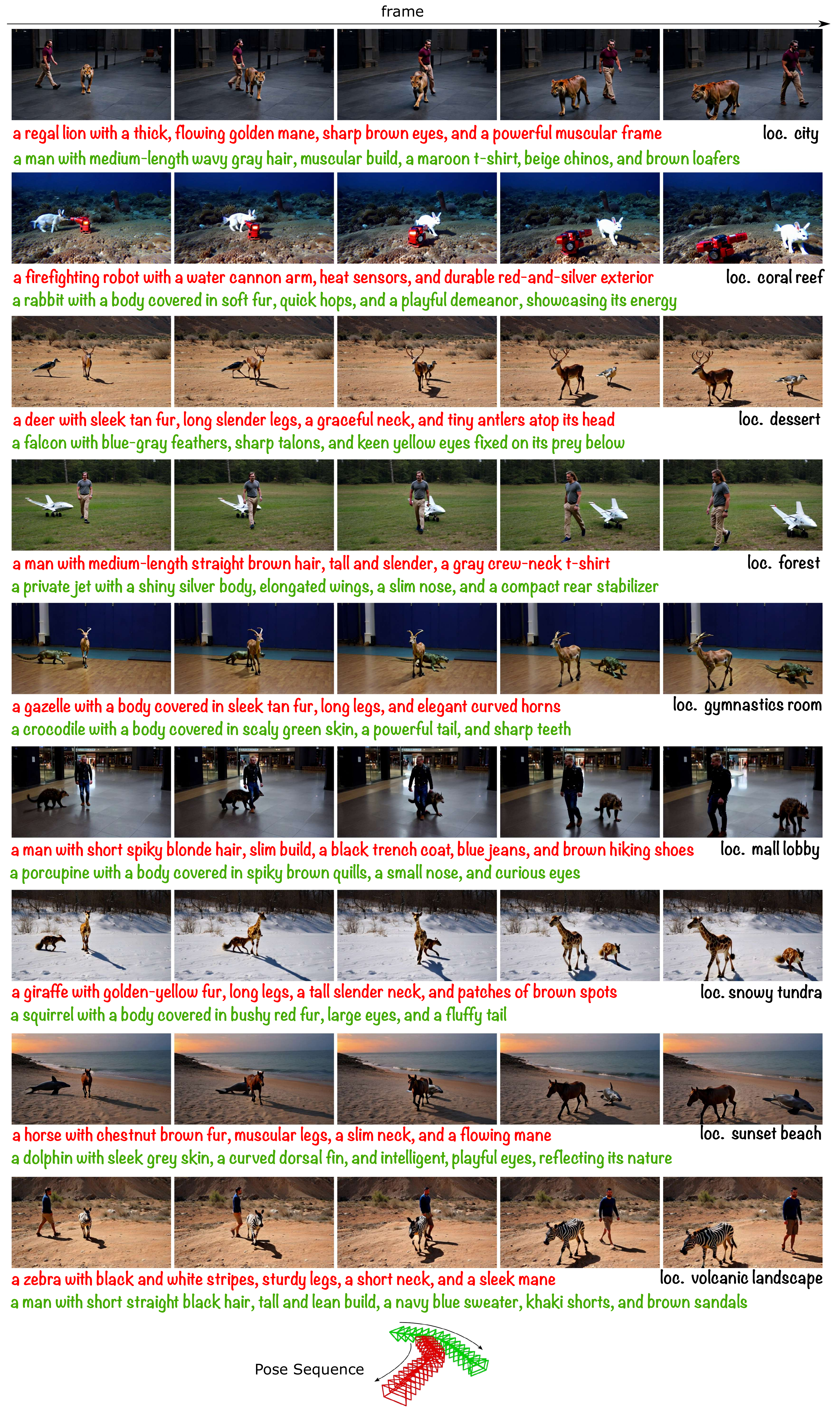}
\caption{\textbf{Generalizable Results with Novel 3D Trajectories \& Entity Prompts (11/20)}}
\end{figure*}

\begin{figure*}[!ht]
\centering
\includegraphics[width=.93\linewidth]{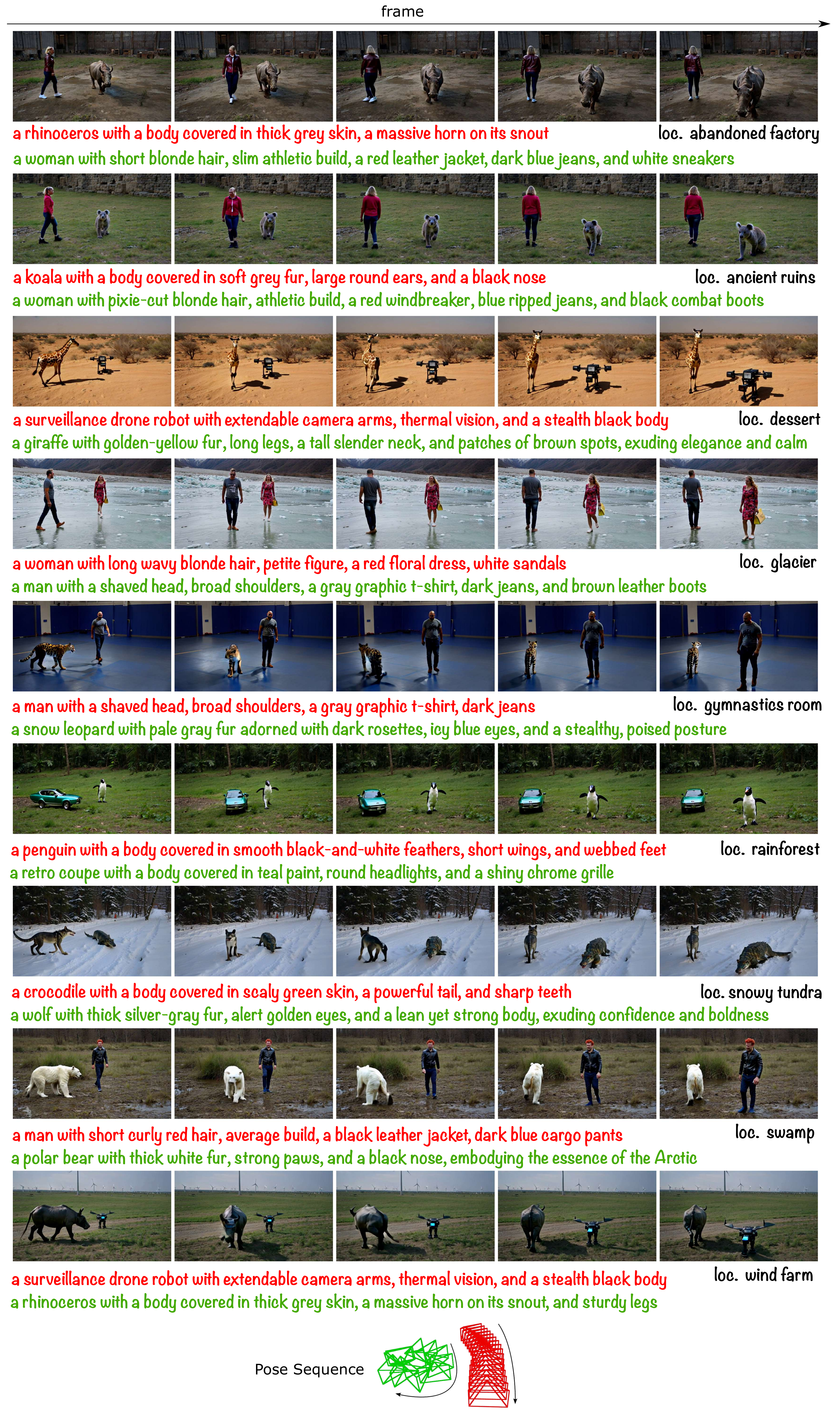}
\caption{\textbf{Generalizable Results with Novel 3D Trajectories \& Entity Prompts (12/20)}}
\end{figure*}

\begin{figure*}[!ht]
\centering
\includegraphics[width=.93\linewidth]{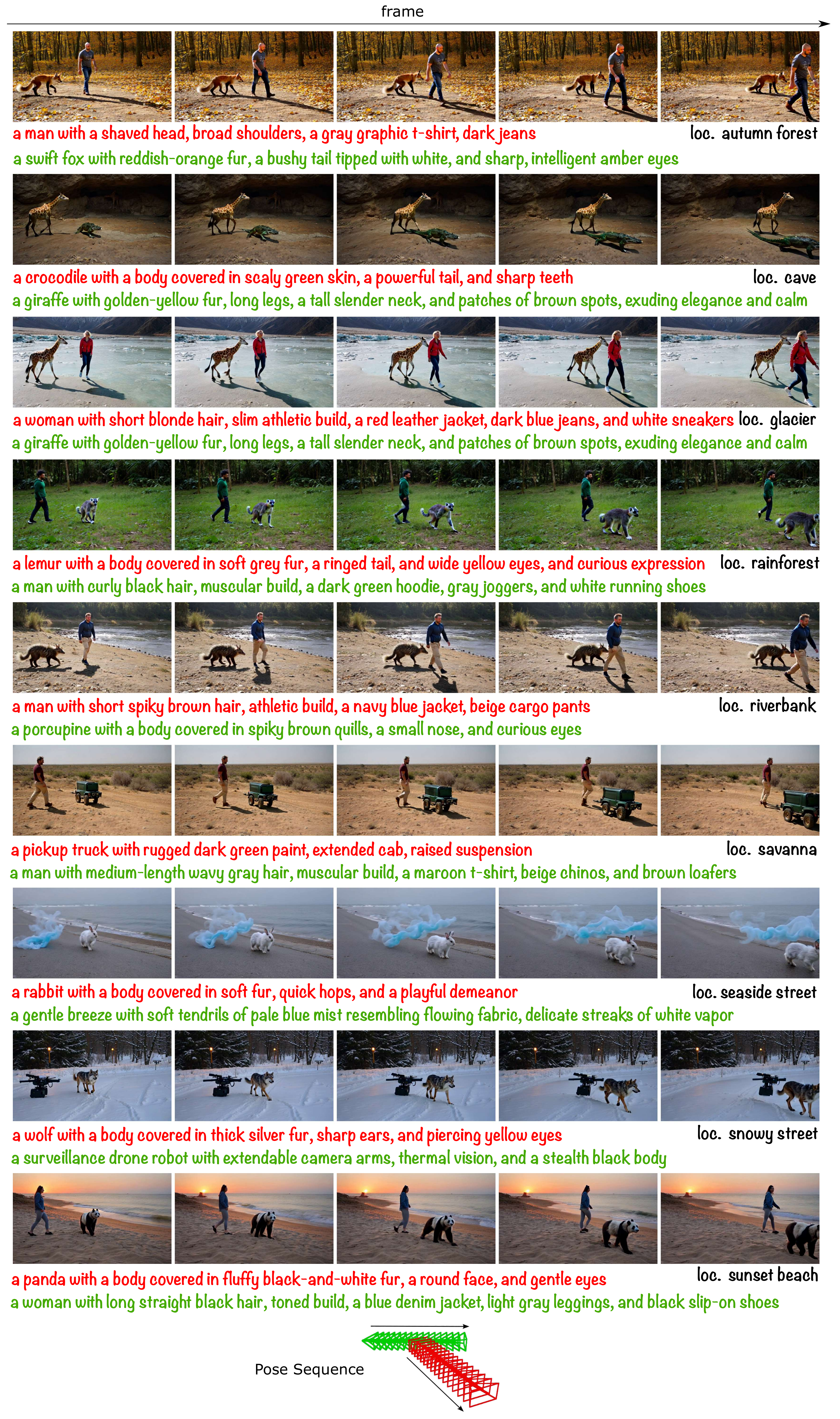}
\caption{\textbf{Generalizable Results with Novel 3D Trajectories \& Entity Prompts (13/20)}}
\end{figure*}

\begin{figure*}[!ht]
\centering
\includegraphics[width=.93\linewidth]{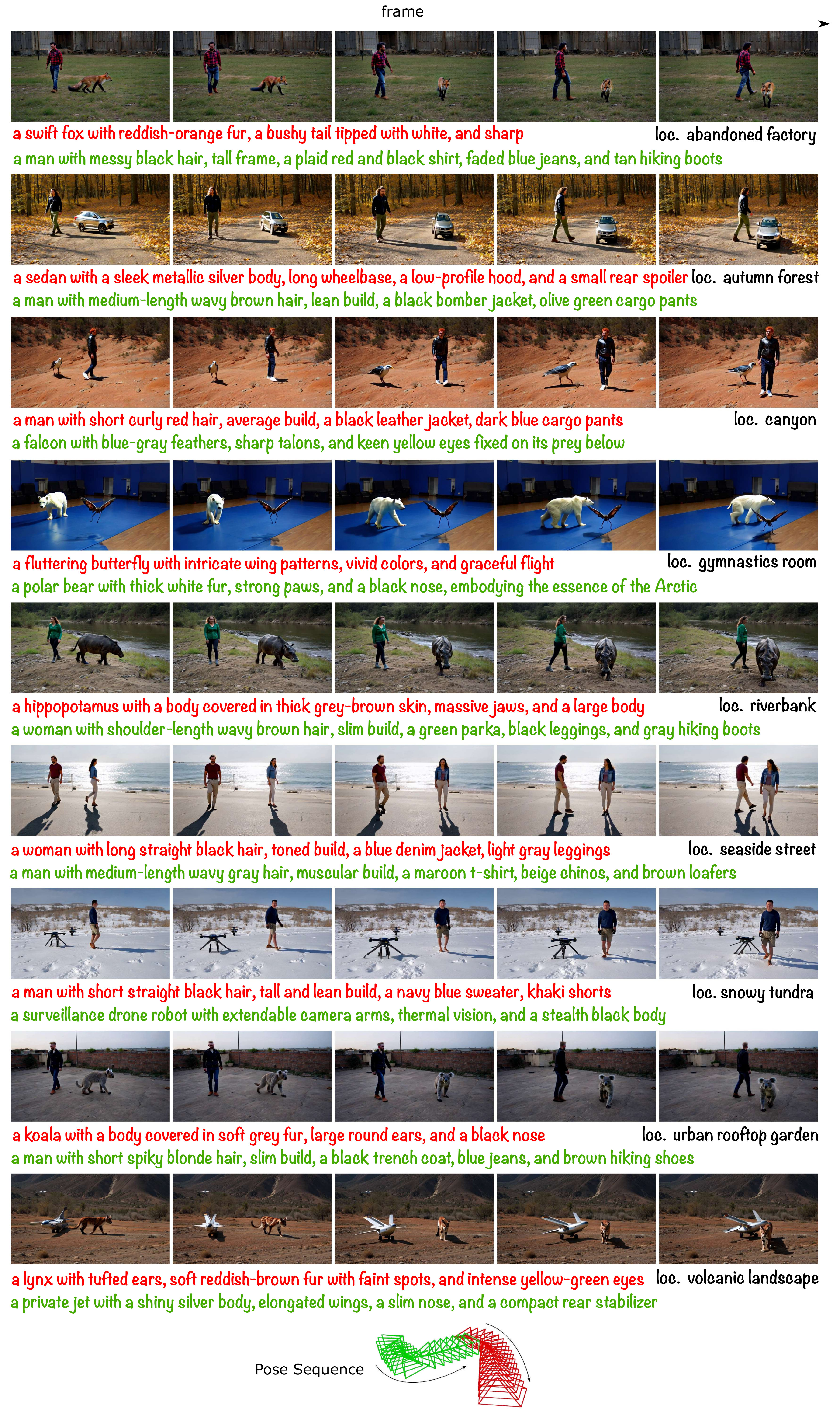}
\caption{\textbf{Generalizable Results with Novel 3D Trajectories \& Entity Prompts (14/20)}}
\end{figure*}

\begin{figure*}[!ht]
\centering
\includegraphics[width=.93\linewidth]{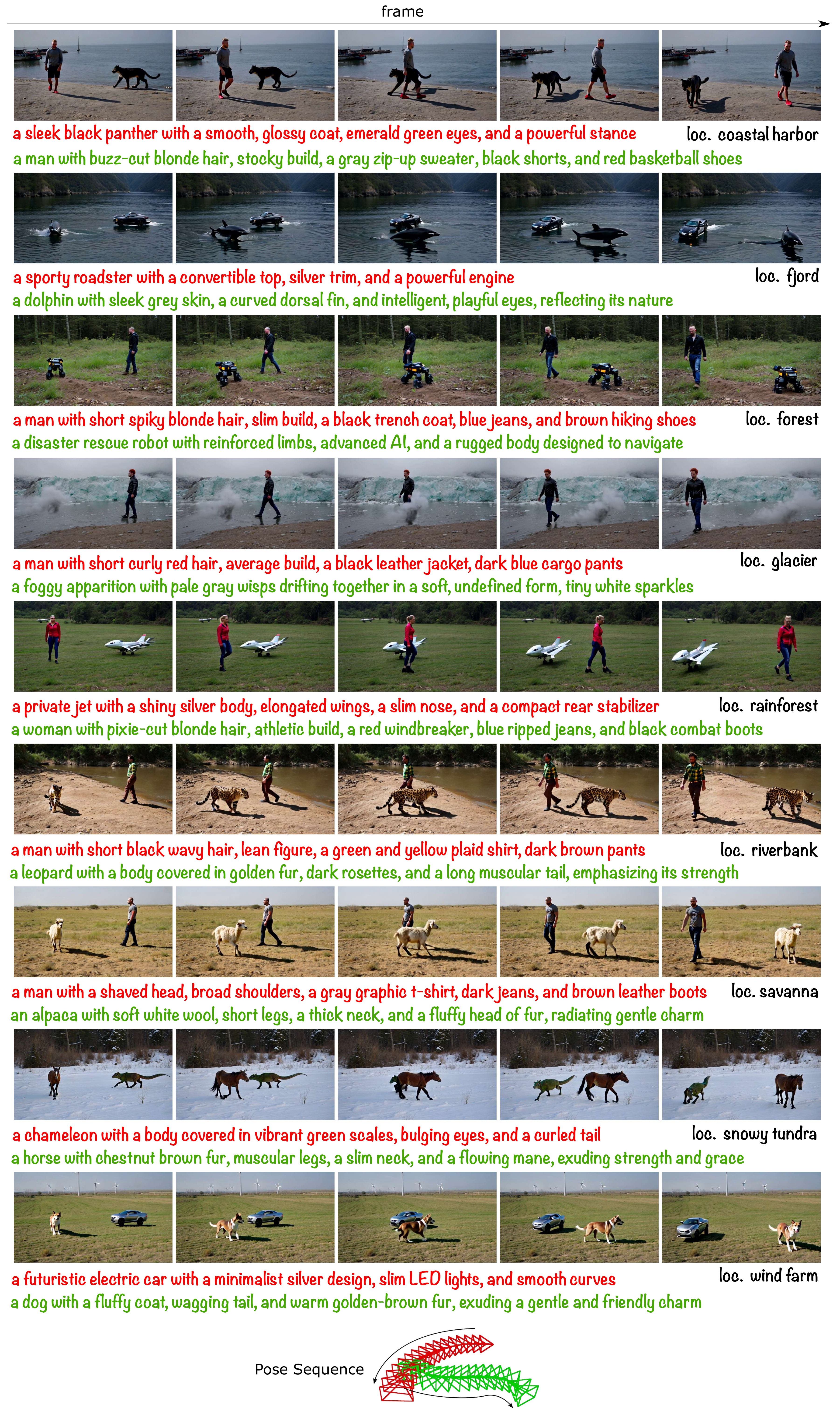}
\caption{\textbf{Generalizable Results with Novel 3D Trajectories \& Entity Prompts (15/20)}}
\end{figure*}

\begin{figure*}[!ht]
\centering
\includegraphics[width=.93\linewidth]{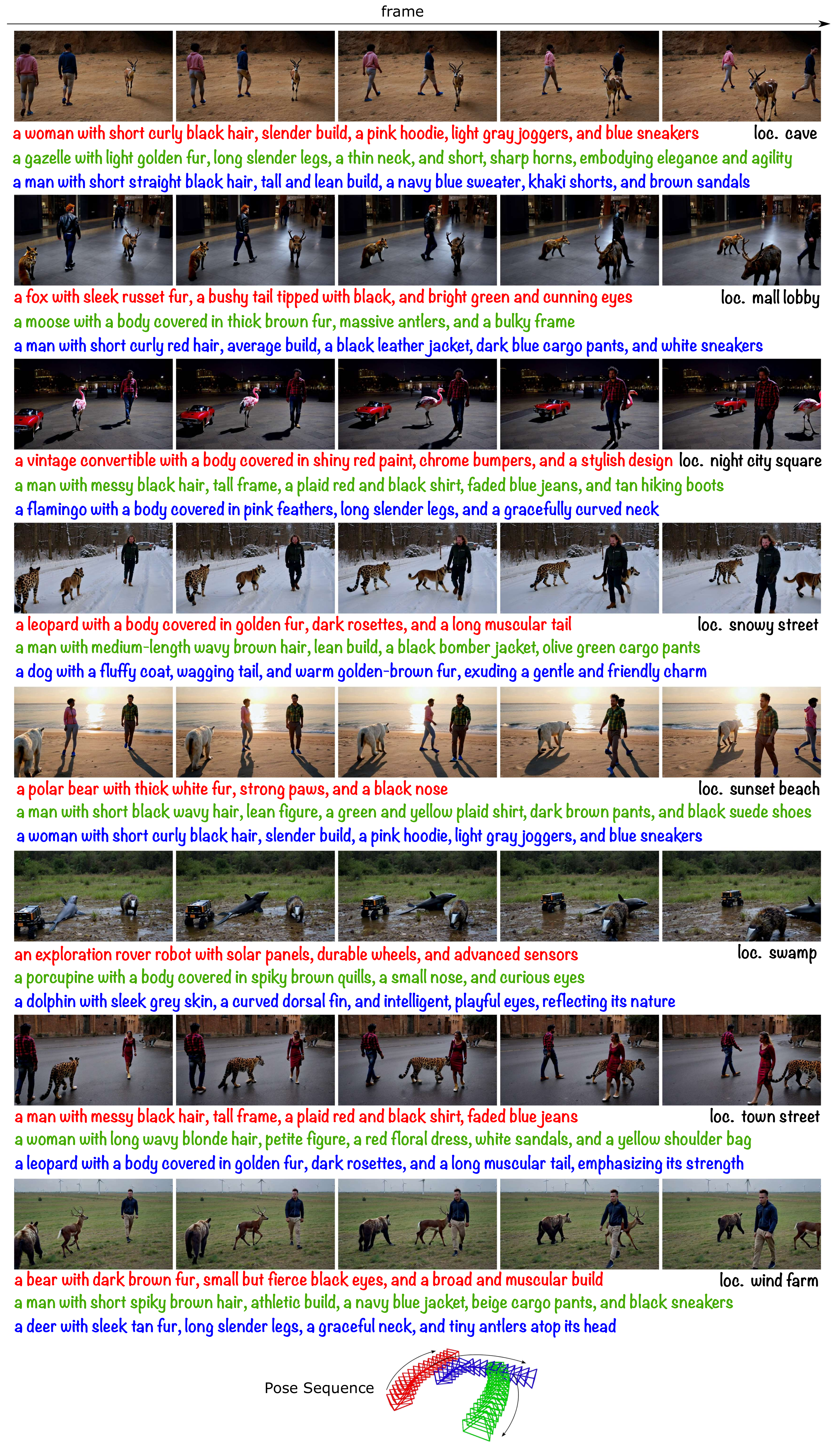}
\caption{\textbf{Generalizable Results with Novel 3D Trajectories \& Entity Prompts (16/20)}}
\end{figure*}

\begin{figure*}[!ht]
\centering
\includegraphics[width=.93\linewidth]{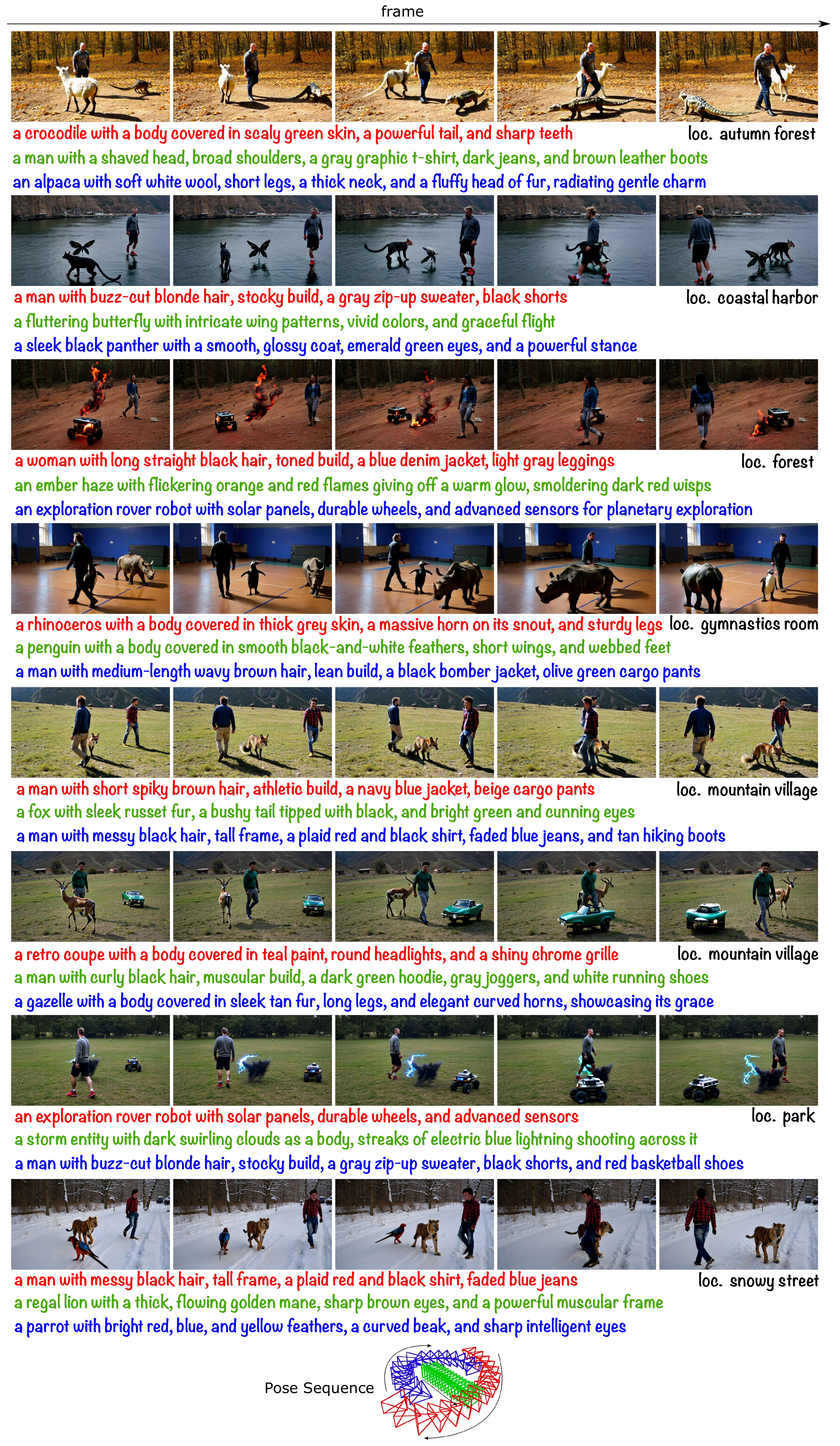}
\caption{\textbf{Generalizable Results with Novel 3D Trajectories \& Entity Prompts (17/20)}}
\end{figure*}

\begin{figure*}[!ht]
\centering
\includegraphics[width=.93\linewidth]{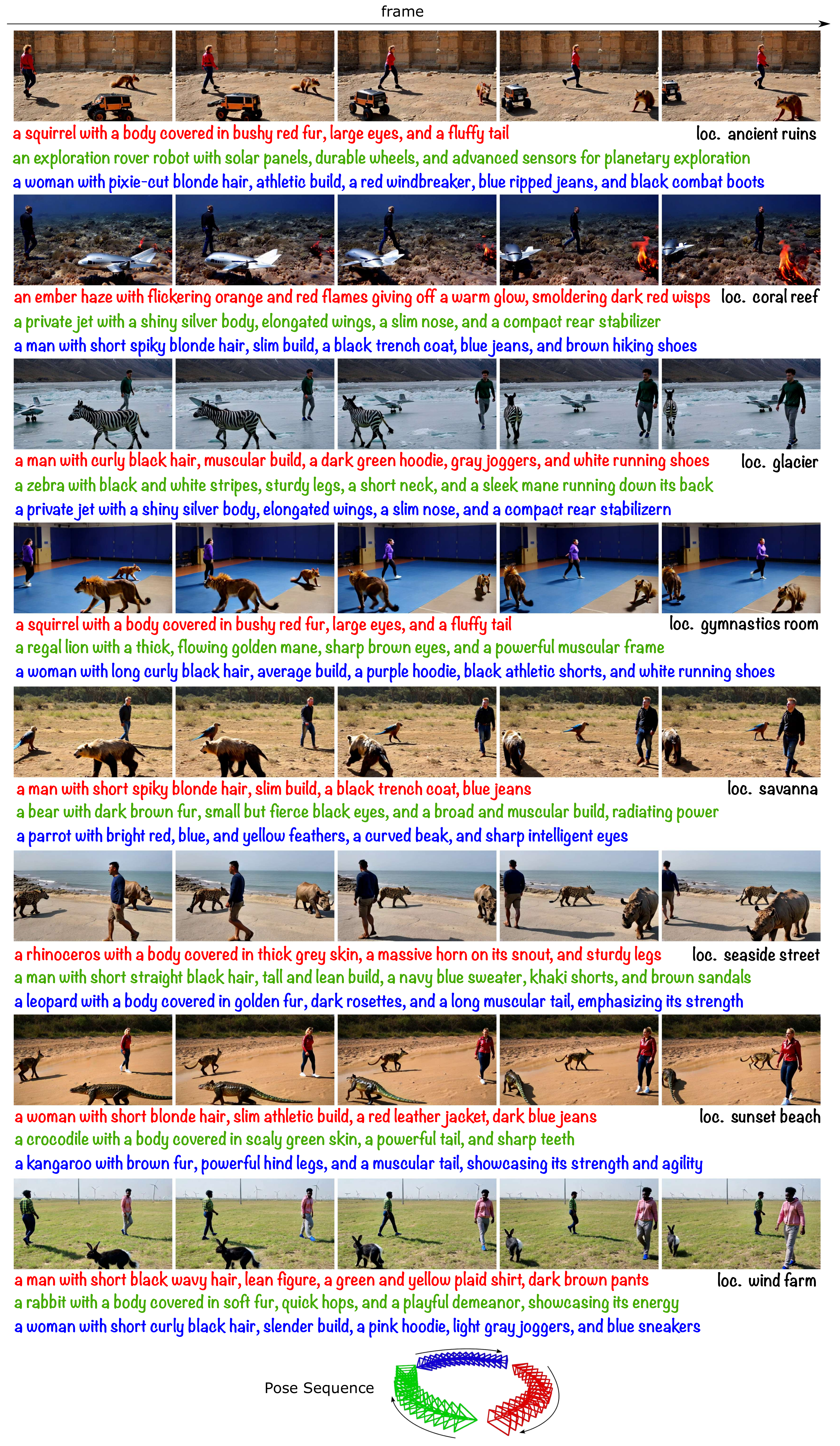}
\caption{\textbf{Generalizable Results with Novel 3D Trajectories \& Entity Prompts (18/20)}}
\end{figure*}

\begin{figure*}[!ht]
\centering
\includegraphics[width=.93\linewidth]{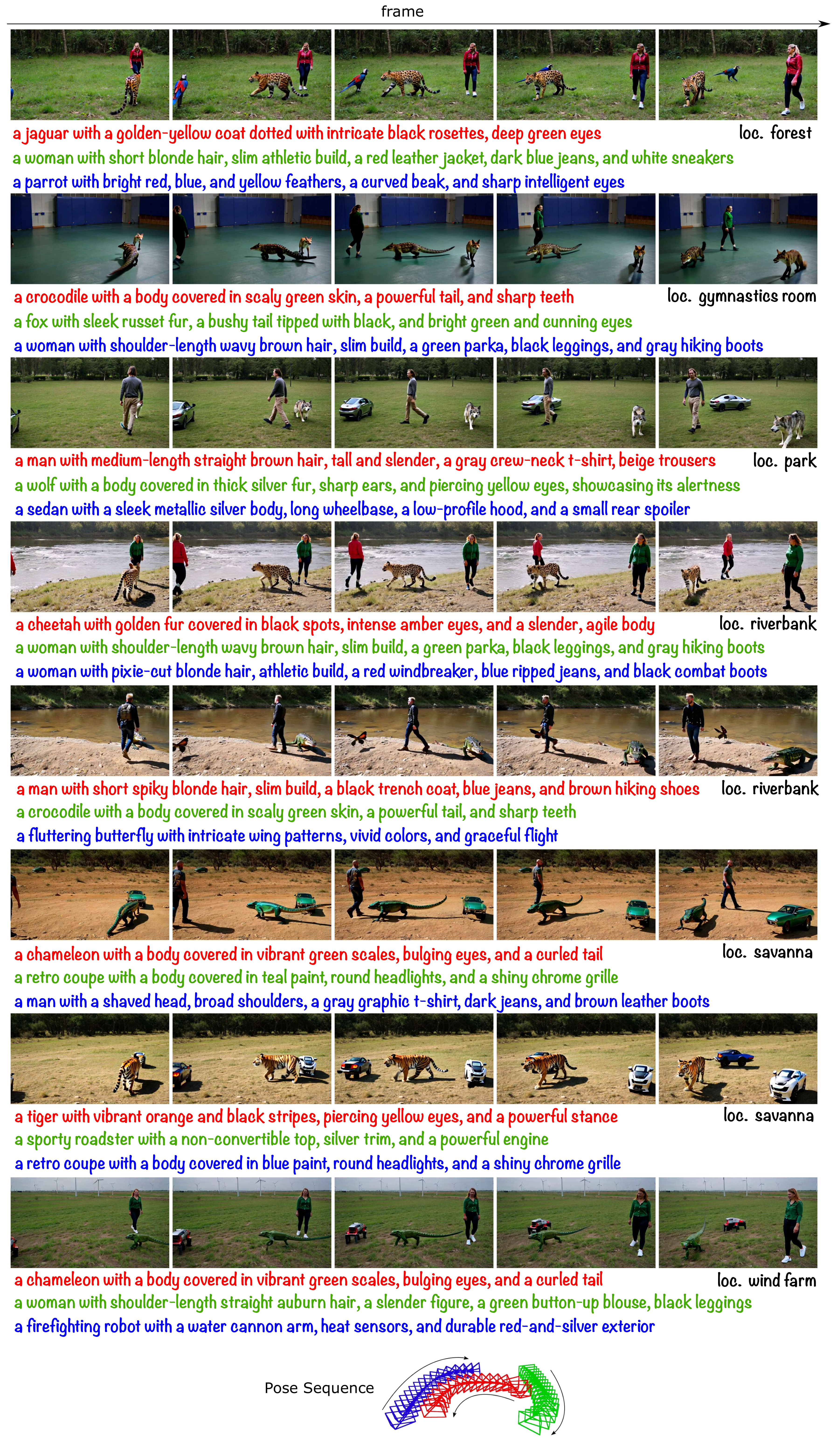}
\caption{\textbf{Generalizable Results with Novel 3D Trajectories \& Entity Prompts (19/20)}}
\end{figure*}

%% file: tab/supp_gptprompt_human.tex
\begin{table}[!ht]

\caption{\textbf{Evaluation Human Prompts}. They are generated using GPT prompt: \textit{``Generate more human samples similar to \{Train Human Sample\}, no more than 25 words.''}}
\label{tab:human_prompts}

\begin{threeparttable}
\resizebox{.99\textwidth}{!} 
{
\begin{tabular}{l}
\toprule

1. a man with short spiky brown hair, athletic build, a navy blue jacket, beige cargo pants, \\ \quad and black sneakers  \\
2. a woman with long wavy blonde hair, petite figure, a red floral dress, white sandals, and a \\ \quad yellow shoulder bag \\
3. a man with a shaved head, broad shoulders, a gray graphic t-shirt, dark jeans, and brown \\ \quad leather boots \\
4. a woman with shoulder-length straight auburn hair, a slender figure, a green button-up \\ \quad 
 blouse, black leggings, and white sneakers \\
5. a man with messy black hair, tall frame, a plaid red and black shirt, faded blue jeans, and \\ \quad tan hiking boots \\
6. a man with medium-length straight brown hair, tall and slender, a gray crew-neck t-shirt,\\ \quad beige trousers, and dark green sneakers \\
7. a woman with short curly black hair, slender build, a pink hoodie, light gray joggers, and \\ \quad blue sneakers \\
8. a man with short black wavy hair, lean figure, a green and yellow plaid shirt, dark brown \\ \quad pants, and black suede shoes \\
9. a man with curly black hair, muscular build, a dark green hoodie, gray joggers, and white \\ \quad running shoes \\
10. a woman with short blonde hair, slim athletic build, a red leather jacket, dark blue jeans, \\ \quad and white sneakers \\
11. a man with medium-length wavy brown hair, lean build, a black bomber jacket, olive \\ \quad green cargo pants, and brown hiking boots \\
12. a man with buzz-cut blonde hair, stocky build, a gray zip-up sweater, black shorts, and red \\ \quad basketball shoes \\
13. a woman with long straight black hair, toned build, a blue denim jacket, light gray legg\\ \quad -ings, and black slip-on shoes \\
14. a man with short curly red hair, average build, a black leather jacket, dark blue cargo pants, \\ \quad and white sneakers \\
15. a woman with shoulder-length wavy brown hair, slim build, a green parka, black leggings, \\ \quad and gray hiking boots \\
16. a man with short straight black hair, tall and lean build, a navy blue sweater, khaki shorts, \\ \quad and brown sandals \\
17. a woman with pixie-cut blonde hair, athletic build, a red windbreaker, blue ripped jeans, \\ \quad and black combat boots \\
18. a man with medium-length wavy gray hair, muscular build, a maroon t-shirt, beige chinos, \\ \quad and brown loafers \\
19. a woman with long curly black hair, average build, a purple hoodie, black athletic shorts, \\ \quad and white running shoes \\
20. a man with short spiky blonde hair, slim build, a black trench coat, blue jeans, and brown \\ \quad hiking shoes \\

\bottomrule

\end{tabular}
}

\end{threeparttable}
\end{table}

%% file: tab/supp_gptprompt_nonhuman_1.tex
\begin{table}[!ht]

\caption{\textbf{Evaluation Non-Human Prompts (1/2)}. They are generated using GPT prompt: \textit{``Generate more animal/car/robot samples similar to \{Train Sample\}, no more than 25 words.''}}
\label{tab:nonhuman_prompts_1}

\begin{threeparttable}
\resizebox{.99\textwidth}{!} 
{
\begin{tabular}{l}
\toprule
1. a dog with a fluffy coat, wagging tail, and warm golden-brown fur, exuding a gentle and \\ \quad friendly charm \\
2. a tiger with vibrant orange and black stripes, piercing yellow eyes, and a powerful stance, \\ \quad exuding strength and grace \\
3. a giraffe with golden-yellow fur, long legs, a tall slender neck, and patches of brown spots, \\ \quad exuding elegance and calm \\
4. an alpaca with soft white wool, short legs, a thick neck, and a fluffy head of fur, radiating \\ \quad gentle charm \\
5. a zebra with black and white stripes, sturdy legs, a short neck, and a sleek mane running \\ \quad down its back \\
6. a deer with sleek tan fur, long slender legs, a graceful neck, and tiny antlers atop its head \\
7. a gazelle with light golden fur, long slender legs, a thin neck, and short, sharp horns, \\ \quad embodying elegance and agility \\
8. a horse with chestnut brown fur, muscular legs, a slim neck, and a flowing mane, exuding \\ \quad strength and grace \\
9. a sleek black panther with a smooth, glossy coat, emerald green eyes, and a powerful stance \\
10. a cheetah with golden fur covered in black spots, intense amber eyes, and a slender, \\ \quad agile body \\
11. a regal lion with a thick, flowing golden mane, sharp brown eyes, and a powerful muscular \\ \quad frame \\
12. a snow leopard with pale gray fur adorned with dark rosettes, icy blue eyes, and a stealthy, \\ \quad poised posture \\
13. a jaguar with a golden-yellow coat dotted with intricate black rosettes, deep green eyes, \\ \quad and a muscular build \\
14. a wolf with thick silver-gray fur, alert golden eyes, and a lean yet strong body, exuding \\ \quad confidence and boldness \\
15. a tiger with a pristine white coat marked by bold black stripes, bright blue eyes, and a \\ \quad graceful, poised form \\
16. a lynx with tufted ears, soft reddish-brown fur with faint spots, and intense yellow-green \\ \quad eyes \\
17. a bear with dark brown fur, small but fierce black eyes, and a broad and muscular build, \\ \quad radiating power \\
18. a swift fox with reddish-orange fur, a bushy tail tipped with white, and sharp, intelligent \\ \quad amber eyes \\
19. a falcon with blue-gray feathers, sharp talons, and keen yellow eyes fixed on its prey below \\
20. a fox with sleek russet fur, a bushy tail tipped with black, and bright green and cunning eyes \\
21. a kangaroo with brown fur, powerful hind legs, and a muscular tail, showcasing its strength \\ \quad and agility \\
22. a polar bear with thick white fur, strong paws, and a black nose, embodying the essence \\ \quad of the Arctic \\
23. a cheetah with a slender build, spotted golden fur, and sharp eyes, epitomizing speed and \\ \quad agility \\
24. a dolphin with sleek grey skin, a curved dorsal fin, and intelligent, playful eyes, reflecting \\ \quad its nature \\
25. a wolf with a body covered in thick silver fur, sharp ears, and piercing yellow eyes, \\ \quad showcasing its alertness \\
26. a leopard with a body covered in golden fur, dark rosettes, and a long muscular tail, \\ \quad emphasizing its strength \\
27. a penguin with a body covered in smooth black-and-white feathers, short wings, and \\ \quad webbed feet \\
28. a gazelle with a body covered in sleek tan fur, long legs, and elegant curved horns, \\showcasing its grace \\

\bottomrule

\end{tabular}
}

\end{threeparttable}
\end{table}

%% file: tab/supp_gptprompt_nonhuman_2.tex
\begin{table}[!ht]

\caption{\textbf{Evaluation Non-Human Prompts (2/2)}. They are generated using GPT prompt: \textit{``Generate more animal/car/robot samples similar to \{Train Sample\}, no more than 25 words.''}}
\label{tab:nonhuman_prompts_2}

\begin{threeparttable}
\resizebox{.99\textwidth}{!} 
{
\begin{tabular}{l}
\toprule
29. a rabbit with a body covered in soft fur, quick hops, and a playful demeanor, showcasing \\ \quad its energy \\
30. a koala with a body covered in soft grey fur, large round ears, and a black nose, radiating \\ \quad cuteness \\
31. a rhinoceros with a body covered in thick grey skin, a massive horn on its snout, and \\ \quad sturdy legs \\
32. a flamingo with a body covered in pink feathers, long slender legs, and a gracefully \\ \quad curved neck \\
33. a parrot with bright red, blue, and yellow feathers, a curved beak, and sharp eyes \\
34. a hippopotamus with a body covered in thick grey-brown skin, massive jaws, and a \\ \quad large body \\
35. a crocodile with a body covered in scaly green skin, a powerful tail, and sharp teeth \\
36. a moose with a body covered in thick brown fur, massive antlers, and a bulky frame \\
37. a fluttering butterfly with intricate wing patterns, vivid colors, and graceful flight \\
38. a chameleon with a body covered in vibrant green scales, bulging eyes, and a curled tail, \\ \quad showcasing its unique charm \\
39. a lemur with a body covered in soft grey fur, a ringed tail, and wide yellow eyes, and \\ \quad curious expression \\
40. a squirrel with a body covered in bushy red fur, large eyes, and a fluffy tail \\
41. a panda with a body covered in fluffy black-and-white fur, a round face, and gentle eyes, \\ \quad radiating warmth \\
42. a porcupine with a body covered in spiky brown quills, a small nose, and curious eyes \\
43. a sedan with a sleek metallic silver body, long wheelbase, a low-profile hood, and a small \\ \quad rear spoiler \\
44. an SUV with a matte black exterior, elevated suspension, a tall roofline, and a compact \\ \quad rear roof rack \\
45. a pickup truck with rugged dark green paint, extended cab, raised suspension, and a modest \\ \quad cargo bed cover \\
46. a vintage convertible with a body covered in shiny red paint, chrome bumpers, and a stylish \\ \quad design \\
47. a futuristic electric car with a minimalist silver design, slim LED lights, and smooth curves \\
48. a compact electric vehicle with a silver finish, aerodynamic profile, and efficient battery \\
49. a firefighting robot with a water cannon arm, heat sensors, and durable red-and-silver exterior \\
50. an industrial welding robot with articulated arms, a laser precision welder, and heat-resistant \\ \quad shields \\
51. a disaster rescue robot with reinforced limbs, advanced AI, and a rugged body designed \\ \quad to navigate \\
52. an exploration rover robot with solar panels, durable wheels, and advanced sensors for \\ \quad planetary exploration \\

\bottomrule

\end{tabular}
}

\end{threeparttable}
\end{table}

%% file: tab/supp_gptprompt_location.tex
\begin{table}[!ht]

\caption{\textbf{Evaluation Location Prompts}. }
\label{tab:location_prompts}

\begin{threeparttable}
\resizebox{.99\textwidth}{!} 
{
\begin{tabular}{l}
\toprule
1. fjord 2. sunset beach 3. cave 4. snowy tundra 5. prairie 6. asian town 7. rainforest 8. canyon\\
9. savanna 10. urban rooftop garden 11. swamp 12. riverbank 13. coral reef 14. volcanic landscape \\
15. wind farm 16. town street 17. night city square 18. mall lobby 19. glacier 20. seaside street 
\\
21. gymnastics room 22. abandoned factory 23. autumn forest 24. mountain village 25. coastal harbor \\
26. ancient ruins 27. modern metropolis 28. dessert 29. forest 30. city 31. snowy street 32. park \\

\bottomrule

\end{tabular}
}

\end{threeparttable}
\end{table}

%% file: tab/ablation_annealed_tc.tex
\begin{table}[!ht]

\caption{Ablation Study on Annealed Timestep $T_c$.}
\label{tab:ablation_annealed_tc}

\centering
\begin{threeparttable}

\resizebox{0.85\textwidth}{!} 
{
\begin{tabular}{ccccccccccc}
\toprule
&  \multicolumn{3}{c}{ Video Quality } & \multicolumn{2}{c}{ 3D Trajectory Accuracy } \\
\cmidrule(lr){2-4} 
\cmidrule(lr){5-6} 
Annealed Timestep $T_c$ & FVD $\downarrow$ & FID $\downarrow$ & CLIPSIM $\uparrow$ & TransErr (m) $\downarrow$ & RotErr (deg) $\downarrow$ \\

\midrule
$T_{c}=5$ & \textbf{1492.79} & \textbf{76.95} & \textbf{0.3469} & 0.844 & 1.099 \\
$T_{c}=10$ & \underline{1976.01} & \underline{106.45} & \underline{0.3429} & 0.546 & 0.493  \\
$T_{c}=15$ & 2179.15 & 122.55 & 0.3405 & 0.437 & 0.422 \\
$T_{c}=20$ & 2236.05 & 128.89 & 0.3374 & 0.391 & 0.284 \\
$T_{c}=25$ & 2240.40 & 132.90 & 0.3337 & \textbf{0.344} & 0.274 \\
$T_{c}=30$ & 2295.13 & 137.52 & 0.3314 & 0.360 & \textbf{0.261} \\
$T_{c}=35$ & 2323.20 & 142.71 & 0.3276 & 0.352 & \underline{0.264} \\
$T_{c}=40$ & 2338.47 & 148.27 & 0.3240 & 0.351 & 0.266 \\
$T_{c}=45$ & 2363.49 & 156.39 & 0.3207 & 0.350 & 0.268 \\
$T_{c}=50$ & 2347.64 & 166.71 & 0.3185 & \underline{0.348} & 0.281 \\

\bottomrule

\end{tabular}
} 
\end{threeparttable}
\end{table}

%% file: tab/ablation_lora_alpha.tex
\begin{table}[!ht]

\caption{Ablation Study on LoRA Scalar $\alpha$.} 
\label{tab:ablation_lora_alpha}

\centering
\begin{threeparttable}
\resizebox{.85\textwidth}{!} 
{
\begin{tabular}{ccccccccccc}
\toprule
&  \multicolumn{3}{c}{ Video Quality } & \multicolumn{2}{c}{ 3D Trajectory Accuracy } \\
\cmidrule(lr){2-4} 
\cmidrule(lr){5-6} 
LoRA Scalar $\alpha$ & FVD $\downarrow$ & FID $\downarrow$ & CLIPSIM $\uparrow$ & TransErr (m) $\downarrow$ & RotErr (deg) $\downarrow$   \\

\midrule
$\alpha=0$ & \textbf{1495.38} & \textbf{80.56} & \textbf{0.3467} & 0.646 & 0.900  \\
$\alpha=0.2$ & \underline{1976.01} & \underline{106.45} & \underline{0.3429}  & 0.546 & 0.493 \\
$\alpha=0.4$ & 2150.42 & 133.76 & 0.3367 & 0.444 & \underline{0.428}   \\
$\alpha=0.6$ & 2330.56 & 152.12 & 0.3277 & 0.394 & \textbf{0.393}  \\
$\alpha=0.8$ & 2318.78 & 195.93 & 0.3125 & \underline{0.378} & 0.450  \\
$\alpha=1.0$ & 2481.33 & 224.81 & 0.3087 & \textbf{0.358} & 0.432 \\
\bottomrule

\end{tabular}
}

\end{threeparttable}
\end{table}

%% file: tab/ablation_training_step.tex
\begin{table}[!ht]

\caption{Ablation Study on Training Step $TS$.}
\label{tab:ablation_ts}

\centering
\begin{threeparttable}
\resizebox{.85\textwidth}{!} 
{
\begin{tabular}{ccccccccccc}
\toprule
&  \multicolumn{3}{c}{ Video Quality } & \multicolumn{2}{c}{ 3D Trajectory Accuracy } \\
\cmidrule(lr){2-4} 
\cmidrule(lr){5-6} 
Train. Steps $TS$ & FVD $\downarrow$ & FID $\downarrow$ & CLIPSIM $\uparrow$ & TransErr (m) $\downarrow$ & RotErr (deg) $\downarrow$ \\

\midrule
$TS=12,000$ & \textbf{1493.68} & \textbf{72.03} & \underline{0.3427} & 0.561 & 0.713 \\
$TS=36,000$ & \underline{1883.15} & \underline{99.98} & 0.3408 & 0.523 & 0.631 \\
$TS=72,000$ & 1976.01 & 106.45 & \textbf{0.3429} & 0.546 & 0.493 \\
$TS=108,000$ & 2068.43 & 111.01 & 0.3388 & \underline{0.446} & \textbf{0.480} \\
$TS=144,000$ & 2102.28 & 114.84 & 0.3367 & \textbf{0.411} & \underline{0.482} \\
\bottomrule

\end{tabular}
}

\end{threeparttable}
\end{table}

%% file: tab/ablation_negative_pose.tex
\begin{table}[!ht]

\caption{Ablation Study on Negative Pose Sequences.}
\label{tab:ablation_neg_pose}

\centering
\begin{threeparttable}
\resizebox{.92\textwidth}{!} 
{
\begin{tabular}{ccccccccccc}
\toprule
&  \multicolumn{3}{c}{ Video Quality } & \multicolumn{2}{c}{ 3D Trajectory Accuracy } \\
\cmidrule(lr){2-4} 
\cmidrule(lr){5-6} 
Negative Condition & FVD $\downarrow$ & FID $\downarrow$ & CLIPSIM $\uparrow$ & TransErr (m) $\downarrow$ & RotErr (deg) $\downarrow$ \\

\midrule
Neg. Pose = Static Motions  & 2141.39 & 118.22 & 0.3360 & \textbf{0.371} & \textbf{0.448} \\
Neg. Pose = Pos. Pose  & \textbf{1976.01} & \textbf{106.45} & \textbf{0.3429} & 0.546 & 0.493 \\
\bottomrule

\end{tabular}
}

\end{threeparttable}
\end{table}